\documentclass{article}

\usepackage{xcolor}
\usepackage[preprint]{corl_2026}
\usepackage[utf8]{inputenc}
\usepackage[T1]{fontenc}
\usepackage{amsmath,amssymb,mathtools}
\usepackage{booktabs}
\usepackage{array}
\usepackage{graphicx}
\usepackage{enumitem}
\usepackage{tikz}
\usetikzlibrary{arrows.meta,positioning,fit,calc}
\usepackage{float}
\usepackage{caption}
\usepackage{overpic}
\usepackage{subcaption}
\usepackage{siunitx}
\usepackage{cleveref}
\crefformat{section}{\S#2#1#3} 
\crefformat{subsection}{\S#2#1#3}
\crefformat{subsubsection}{\S#2#1#3}

\DeclareMathOperator{\clip}{clip}
\newcommand{\R}{\mathbb{R}}

\title{TACT-ful: Multi-Channel \underline{T}errain \underline{A}ffordance and \underline{C}ompliance \underline{T}raining for Payload-Robust Perceptive Humanoid Locomotion}

\author{
  \mdseries
  Thanh Ly$^{1}$\thanks{Equal contribution.\enspace
    Corresponding author: An T.~Le, \texttt{an@robot-learning.de}.}\quad
  Truong-Duy Dang$^{1}$\footnotemark[1]\quad
  Chien Le$^{1}$\quad
  Tan-Dzung Do$^{1}$\quad
  Phuong Tuan Dat$^{1}$\\[3pt]
  Cuc T.~Trinh$^{1}$\quad
  Vien Anh Ngo$^{1, 2}$\quad
  An T. Le$^{1, 2, 3}$\\[6pt]
  $^{1}$VinRobotics, Vietnam\quad
  $^{2}$Center for AI Research, VinUniversity, Vietnam\\[1pt]
  $^{3}$Intelligent Autonomous Systems, TU Darmstadt, Germany
}

\begin{document}
\maketitle

\vspace{-7mm}
\begin{figure}[h!]
  \centering
  \begin{subfigure}{1.0\textwidth}
    \includegraphics[width=\textwidth]{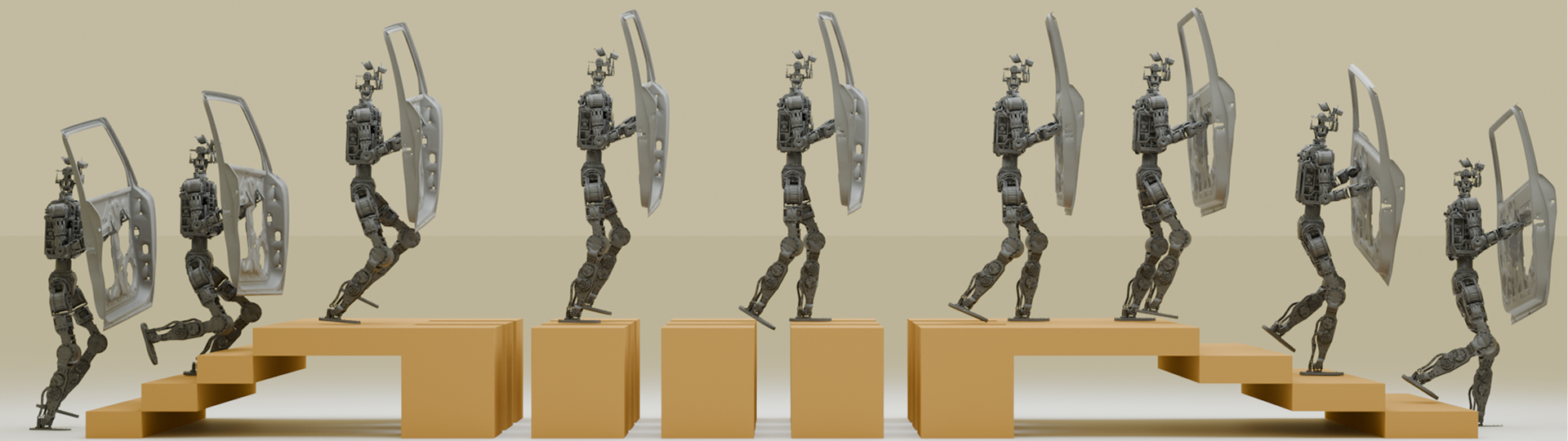}
  \end{subfigure} \\%
  \vspace{-0.3ex}
  \begin{subfigure}{0.2\textwidth}
    \includegraphics[width=\textwidth]{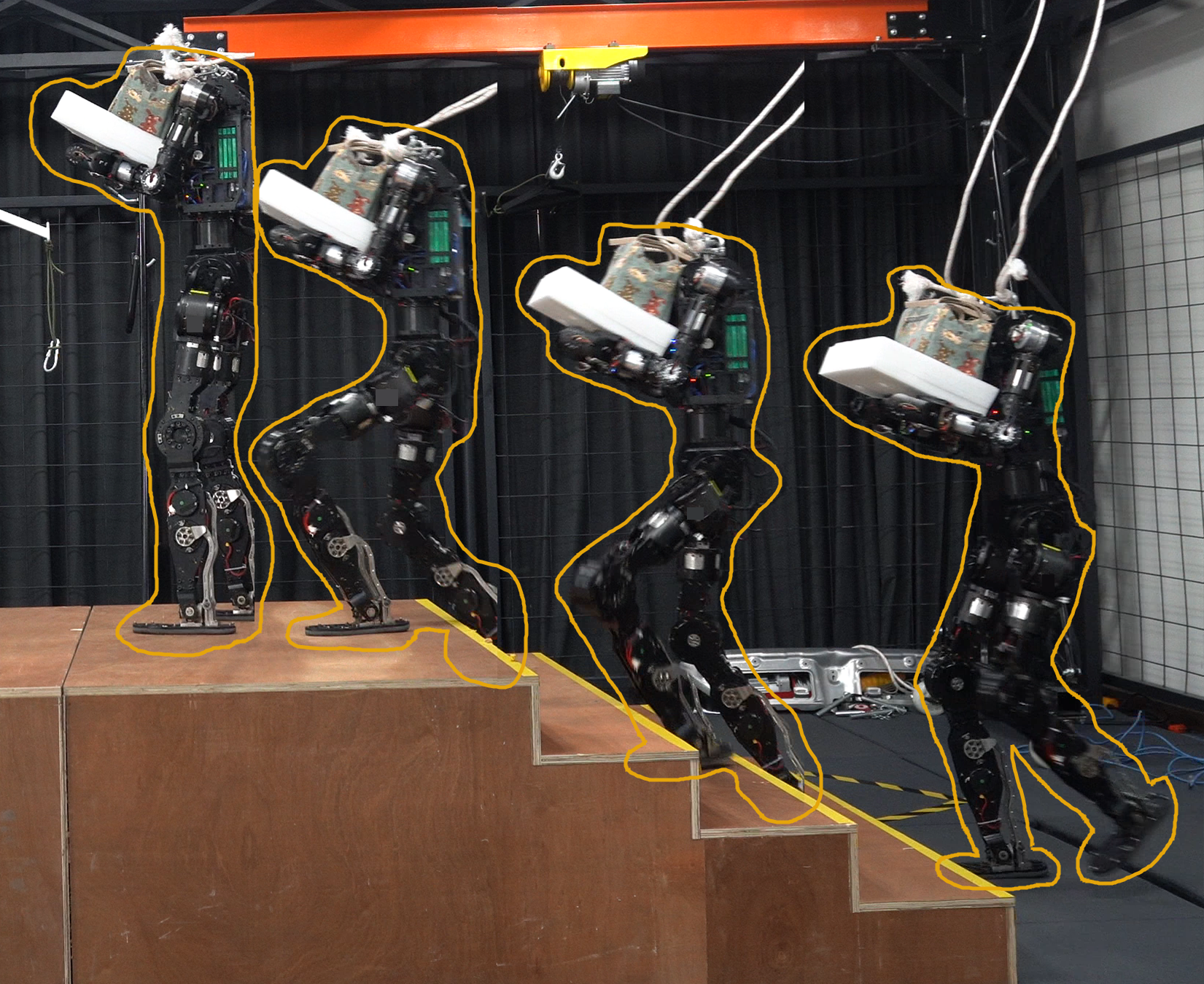}
  \end{subfigure}%
  \begin{subfigure}{0.2\textwidth}
    \includegraphics[width=\textwidth]{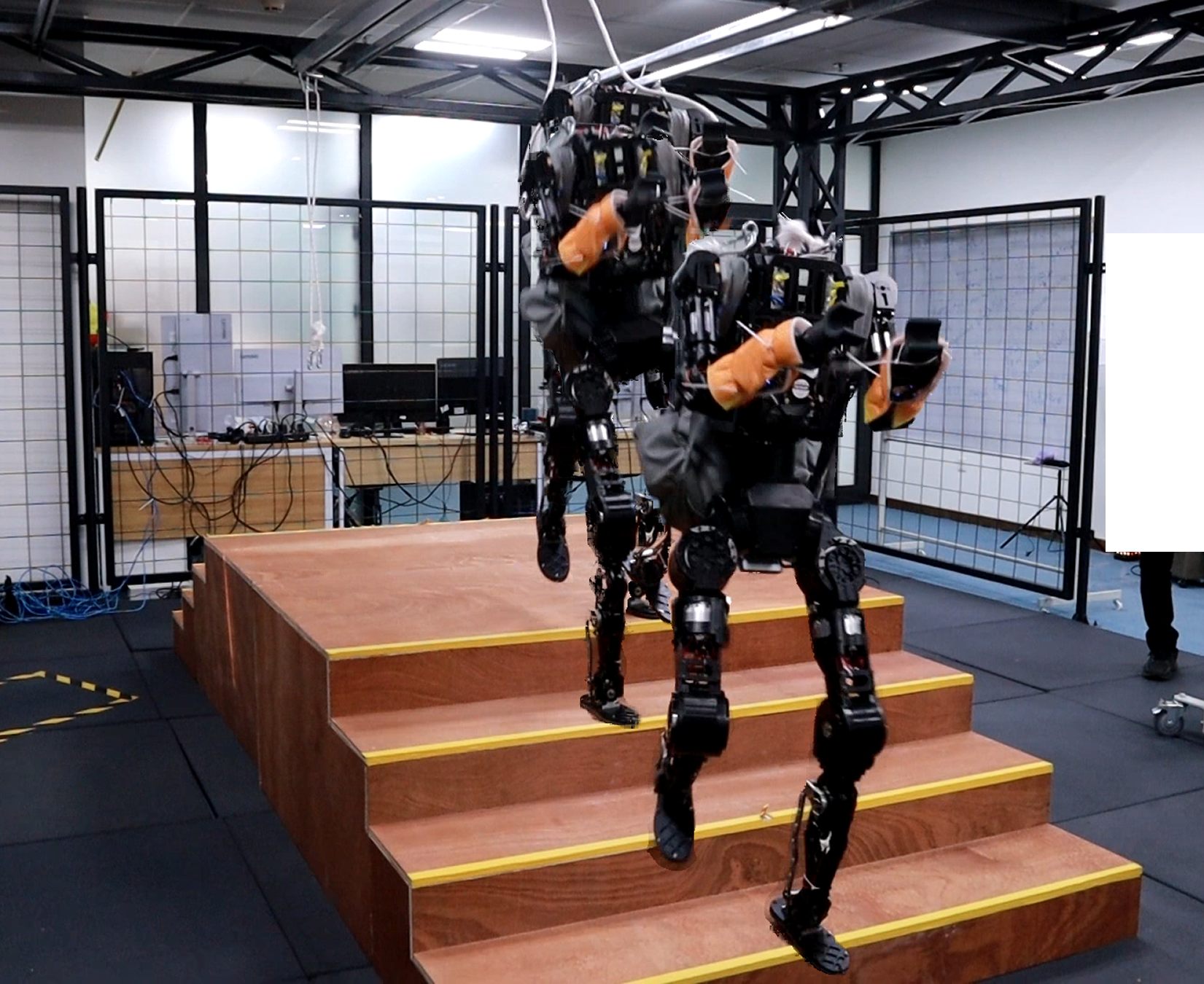}
  \end{subfigure}%
  \begin{subfigure}{0.2\textwidth}
    \includegraphics[width=\textwidth]{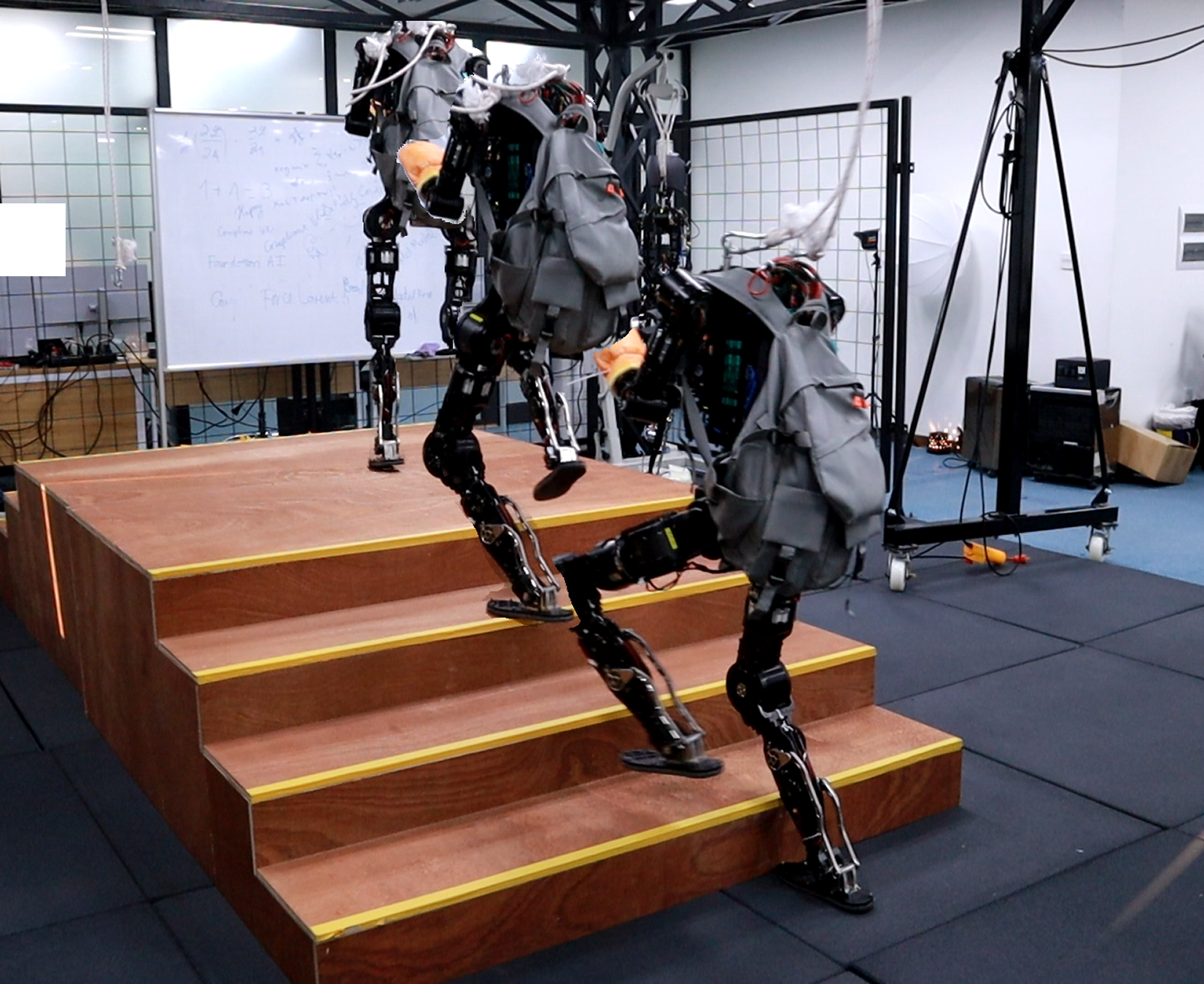}
  \end{subfigure}%
  \begin{subfigure}{0.2\textwidth}
    \includegraphics[width=\textwidth]{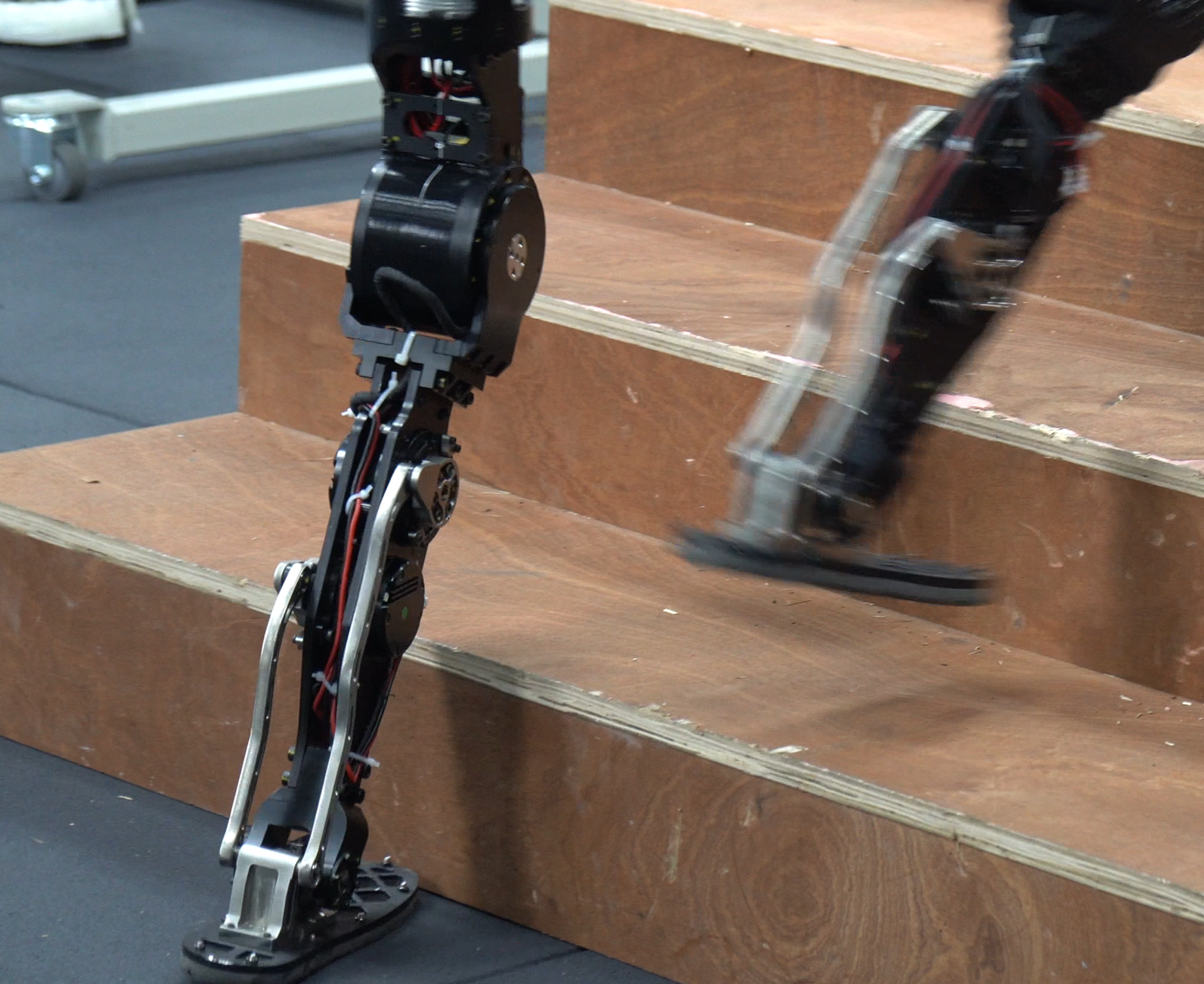}
  \end{subfigure}%
  \begin{subfigure}{0.2\textwidth}
    \includegraphics[width=\textwidth]{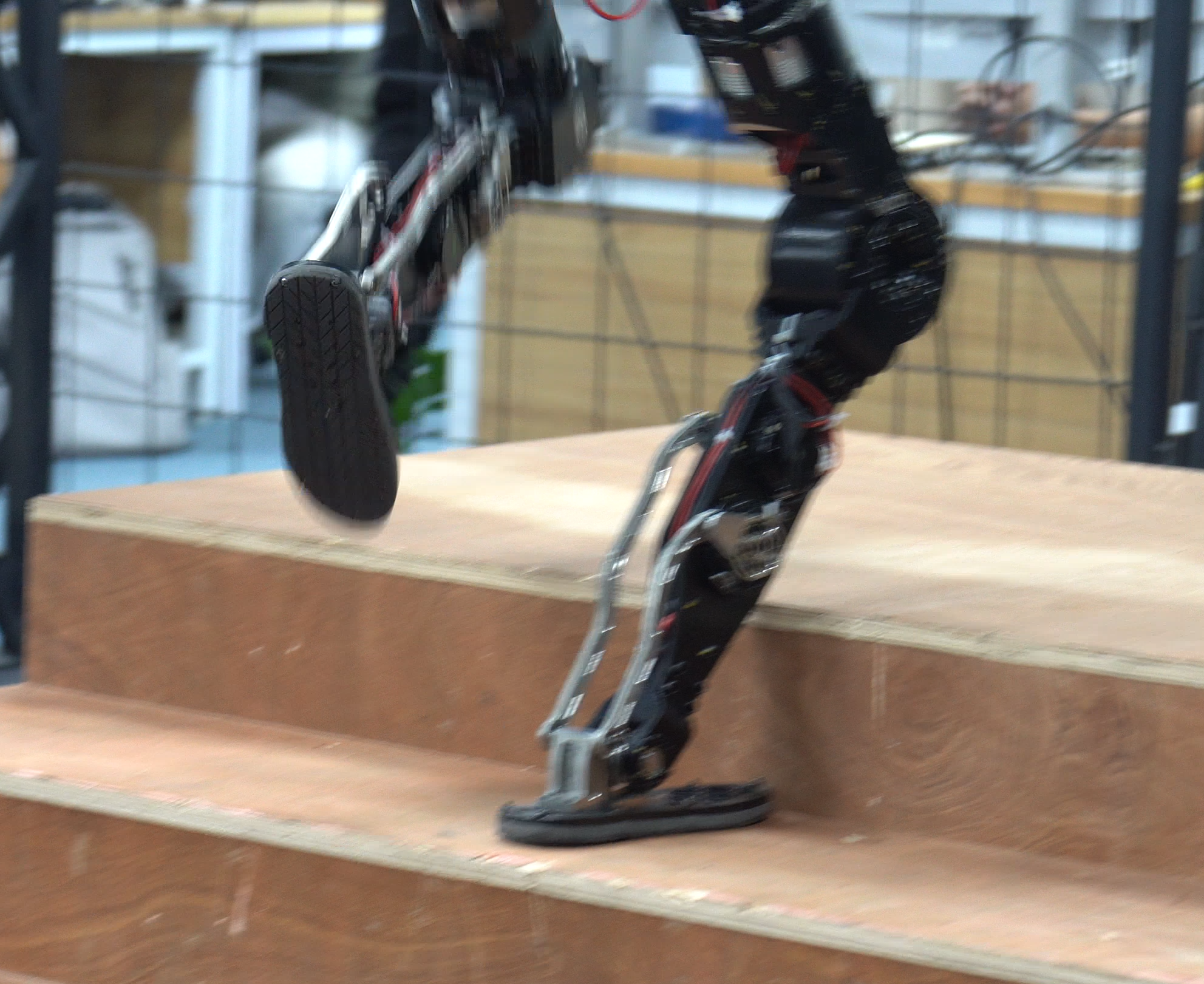}
  \end{subfigure}%
  \caption{A service humanoid, trained on TACT-ful, traverses structured terrain while carrying
    heavy payload.}%
  \label{fig:teaser}
\end{figure}

\vspace{-3mm}
\begin{abstract}
  Foothold selection on structured terrain requires explicit reasoning about contact planarity, surface steepness, and kinematic reachability, properties not captured by a single height-based terrain signal.
  We propose a multi-channel terrain cost combining flatness, steepness, and velocity-aware height feasibility, plus a forward climb reward, that simultaneously drives a GPU-parallel divergent component of motion (DCM) foothold planner and shapes a dense per-step affordance reward for an asymmetric actor-critic policy trained with proximal policy optimization (PPO) from depth images.
  A Bézier swing trajectory with adaptive apex bias extends foothold tracking to joint position-and-orientation, using the arc tangent to guide sole orientation through riser crossings and tread landings.
  To support payload tasks, we introduce a lower-body compliance training procedure in which a virtual wrench is injected at a sampled load attachment point, generating physically consistent force and moment;
  wrench-aware compliance targets replace rigid pose penalties, and the policy learns to yield to load-induced perturbations without force sensing.
  The full system trains end-to-end with standard PPO, no distillation, and no teacher-student staging, and is deployed on a humanoid directly from simulation with configuration changes only.
  In simulation, the policy reaches $1.0~\mathrm{m/s}$ on stairs with risers up to $0.20~\mathrm{m}$ and improves payload robustness up to ${\sim}15~\mathrm{kg}$ centered load and for moment-dominated wrist loads without fine-tuning.
  We also provide a qualitative hardware demonstration on structured terrain.
  Project website: \url{https://fai-rl-tech.github.io/tact-locomotion.github.io/}
\end{abstract}


\section{Introduction}

Service humanoids operating in human environments must traverse structured
non-flat terrain, including staircases, stepped platforms, and ramps, while
carrying tools or cargo. Foothold placement on such terrain is consequential: a
sole that straddles a step edge bears load on a line contact, narrowing the
friction cone and generating lateral impulses that scale with robot mass. The
capture-point~\citep{pratt2006capturepoint} divergence rate is independent of
mass, but the ground reaction force needed to intercept a divergent trajectory
scales with total weight, so that early, terrain-quality-aware foothold
selection~\citep{bostondynamics2020terrain} becomes more consequential as
platform mass grows. Payload carrying adds a second challenge: a carried load
shifts the effective neutral pelvis pose and imposes a time-varying tug on the
torso, requiring the locomotion controller to absorb these perturbations
without loss of balance.

Existing approaches address subsets of this problem. Model-based foothold
planners select discrete landing targets subject to stability and reachability
constraints~\citep{acosta2025perceptive,xiang2026perceptivetiming,kim2025visionfootstep},
but their outputs are not used to shape the reinforcement-learning (RL) policy
reward. Perceptive RL methods for humanoids consume elevation maps or depth
images to adapt gait on rough
terrain~\citep{song2025gaitadaptive,liu2026faststair,ben2025gallant}, but
typically encode terrain through a single height channel and do not reason
explicitly about contact planarity or kinematic overreach. Compliance under
payload is largely treated separately, either through dedicated force
controllers, online payload estimators, or manual gait re-tuning, rather than
as a behavior the locomotion policy acquires within a single terrain-aware
training run.

The contributions of this paper are:
\begin{enumerate}[leftmargin=1.25em,itemsep=2pt,topsep=2pt]
  \item \textbf{Multi-channel terrain cost with terrain-adaptive swing and
          tangent-guided foot orientation.}
        Flatness, steepness, velocity-aware height feasibility, and a climb
        bonus jointly drive a GPU-parallel DCM foothold planner and serve as a
        dense terrain-affordance learning signal for the RL policy.
        A Bézier swing trajectory with adaptive apex bias and clearance extends
        foothold tracking to joint position-and-orientation references, using
        the arc tangent to guide sole orientation through riser crossings and
        tread landings. Unlike the closest predecessor, which couples a
        single-channel DCM planner to a position-only foothold reward under a
        multi-stage curriculum~\citep{liu2026faststair}, our cost adds explicit
        flatness, velocity-aware feasibility, and speed-gated climb channels, and
        the tangent schedule extends the swing reference to sole \emph{orientation}.
  \item \textbf{Lower-body compliance training for payload-robust locomotion.}
        A virtual wrench injected at a sampled load attachment point generates
        physically consistent force and moment; together with wrench-aware compliance
        targets for pelvis height and trunk orientation, it teaches the policy to yield to
        payload-induced perturbations without force sensing, improving robustness up to
        ${\sim}15\,\mathrm{kg}$ centered load and for moment-dominated wrist loads
        without payload-specific fine-tuning.
  \item \textbf{An integrated end-to-end pipeline with a qualitative hardware
          demonstration.}
        The system trains with standard PPO in a single run (no distillation, no
        teacher-student staging) and combines terrain-aware foothold selection
        with payload compliance, and is deployed zero-shot from simulation on a
        service humanoid traversing structured terrain while carrying payload.
\end{enumerate}

\section{Related Work}

\paragraph{Foothold planning and terrain-conditioned reinforcement learning.}
Model-based foothold planners select discrete landing targets under LIPM
capturability constraints: \citet{acosta2025perceptive} formulate MIQP over
convex foothold regions, while \citet{xiang2026perceptivetiming} jointly
optimize step timing and placement under DCM bounds.
\citet{kim2025visionfootstep} replace the MIQP with an RL-trained footstep
policy mapping elevation maps to 3D targets, decoupled from end-to-end
training.\@ \citet{lee2024liprl} use LIP-derived footstep targets as a partial
RL reward, and \citet{liu2026faststair} reformulate DCM foothold search as
GPU-parallel discrete optimization used as an explicit RL reward with a
multi-stage curriculum, the most direct predecessor to our approach.
\citet{wang2025beamdojo} show that a purely learned dense foothold reward with
a double-critic curriculum can traverse sparse footholds without an explicit
planner.

\paragraph{Perceptive humanoid locomotion and terrain representation.}
\citet{agarwal2022leggedlocomotionchallengingterrains,miki2022wild} pioneered
the privileged-training-to-depth-distillation paradigm, decoupling policy and
perception learning for zero-shot real-world deployment, which has become
standard for perceptive legged locomotion.
Building on this, \citet{song2025gaitadaptive} reconstruct a dense egocentric
height map from depth and jointly output joint targets and adaptive gait
frequency; \citet{radosavovic2024learninghumanoidlocomotionchallenging} train a
Transformer policy on diverse outdoor terrain without perceptive reward shaping;
\citet{long2024learninghumanoidlocomotionperceptive} achieve stair climbing via
LiDAR elevation maps; \citet{ben2025gallant} extend perception to full 3D
structure via a voxel-grid representation; \citet{zhang2026rpl} distil
terrain-specific experts into a unified transformer policy with multi-view depth
fusion and velocity-scaled features; and \citet{sun2026nowyousee} train
end-to-end from raw stereo depth images using terrain-specific multi-critic
shaping.
Vision-based whole-body policies demonstrate perceptive parkour, stair
climbing, gap crossing, and platform jumping, on humanoids and
quadrupeds~\citep{zhuang2024humanoidparkourlearning,wu2026php,hoeller2023anymalparkourlearningagile}.
Standard terrain representations include probabilistic elevation
maps~\citep{fankhauser2018probabilistic}, traversability
scoring~\citep{fan2021step}, and multi-layer grid maps~\citep{anyboticsgridmap}.

\paragraph{Payload-carrying and compliant loco-manipulation.}
Large-scale domain randomization yields implicit payload
tolerance~\citep{radosavovic2023realworldhumanoidlocomotionreinforcement};
\citet{kumar2021rma} achieve payload and terrain robustness on quadrupeds
through rapid online system identification; \citet{zhang2025hub} demonstrate
balance recovery under large external perturbations via a multi-phase
curriculum; \citet{fu2026loadawarelocomotioncontrolhumanoid} make payload
awareness explicit on the Tiangong humanoid via a history-based estimator in a
decoupled lower-body RL architecture, evaluated on flat terrain.
\citet{pasricha2025dynamicscomplianttrajectorydiffusionsupernominal} extend
payload limits to $3\times$ nominal via compliant trajectory diffusion. For
force-adaptive compliance without force sensors,
\citet{xu2025facetforceadaptivecontrolimpedance} train legged robots to absorb
collision impulses and sustain payload pulls up to $\tfrac{2}{3}$ body weight
by tracking a virtual impedance reference;
\citet{zhi2025learningunifiedpolicyposition} unify position and force control
for contact-rich loco-manipulation. Terrain property estimation from
vision~\citep{chen2024visionphysical} and friction
identification~\citep{kim2025friction} connect perception to contact mechanics.
None of these works couple terrain-quality-aware foothold selection with
compliance training in a single end-to-end RL formulation, leaving the
interaction between payload mass, step-height feasibility, and terrain
planarity unaddressed.

\section{Terrain Affordance and Compliance Training}%
\label{sec:method}

\begin{figure}[t]
  \centering
  \includegraphics[width=\textwidth]{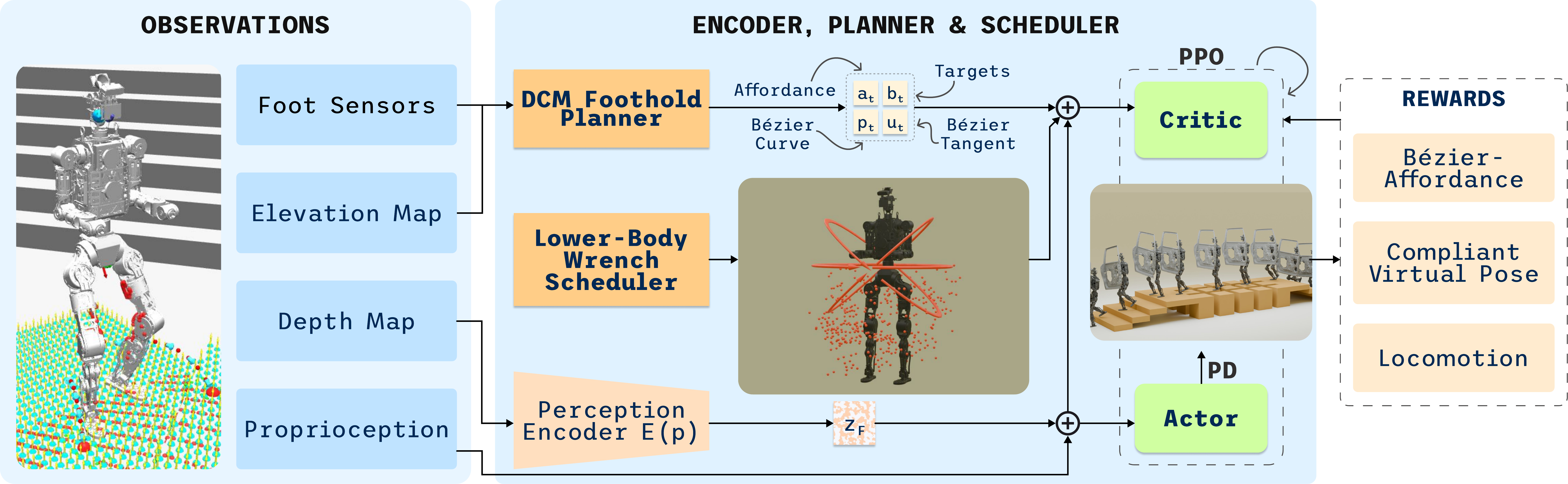}
  \small
  \caption{\textbf{Overview of the proposed framework.} Two parallel modules
    feed into the training reward buffer.\@ \emph{(i) DCM foothold planner:} at
    every control step, a pelvis-mounted elevation map is consumed by the
    GPU-parallel DCM foothold planner, which selects terrain-optimal landing
    targets and produces a B\'{e}zier swing trajectory reference; these targets
    define the foothold-tracking and terrain-specific reward terms and are
    provided as privileged observations to the critic.\@ \emph{(ii) Wrench compliance scheduler:} a virtual wrench is injected at a
    sampled load attachment point, generating force that shifts the pelvis and moment
    that tilts the trunk; wrench-aware compliance targets replace rigid pelvis-height
    and orientation penalties, training the policy to absorb payload-induced
    perturbations without explicit force sensing.}%
  \label{fig:system_overview}
  \vspace{-2mm}
\end{figure}

\Cref{fig:system_overview} Overview the framework: a DCM foothold planner and a
wrench compliance scheduler feed a shared reward for an
asymmetric actor-critic policy. All hyperparameters and implementation details are located in the Appendices.

\subsection{Capture-Point Stability}%
\label{sec:dcm}

We use the divergent component of motion (DCM) / capture-point
framework~\citep{pratt2006capturepoint,englsberger2015dcm} with a constant
natural frequency $\omega_0 = \sqrt{g/z_0}$, where $g$ is gravitational
acceleration and $z_0$ is the nominal center of mass (CoM)
height~\citep{khadiv2020steptiming}. The DCM is defined as $\xi =
  x_{\mathrm{CoM}} + \dot{x}_{\mathrm{CoM}}/\omega_0$ and propagates as $\xi(T) =
  \xi_0\,e^{\omega_0 T}$ over a swing of duration $T$ (step time); applied
component-wise in the horizontal plane this gives the 2D DCM
$\boldsymbol{\xi}_T \in \mathbb{R}^2$. The nominal step offset
$\mathbf{b}_{\mathrm{nom}} = (b_x, b_y)$ keeps the divergent mode in steady
state~\citep{koolen2012capturability}; we adopt the linear-CoM-height reduction
of \citet{liu2026faststair} and set $\omega = \omega_0$ throughout.

\subsection{Multi-Channel Terrain Cost}%
\label{sec:terrain_cost}

The planner selects the cell $i$ that minimizes the total cost
\begin{equation}
  \mathcal{J}_i = \alpha_{\mathrm{pos}}\,d_{\mathrm{pos},i}
  + \alpha_{\mathrm{dcm}}\,d_{\mathrm{dcm},i}
  + \alpha_E\,E_i
  + \alpha_Q\,Q_i
  + \alpha_M\,M_i
  - \alpha_{\mathrm{climb}}\,b_i,
  \label{eq:cost}
\end{equation}
where
$d_{\mathrm{pos},i} = {(x_i - v_x T)}^2 + \beta\,{(y_i - (v_y T + \mathrm{sign}(f)\,l_p))}^2$
is an asymmetric position residual from the nominal stride target
($\beta > 1$ penalizes lateral deviation more than sagittal to limit foot spread;
$\mathbf{p}_i = {(x_i, y_i)}^\top$ is the 2D candidate foothold, $(v_x, v_y)$ the
commanded planar velocity, $l_p$ the nominal lateral inter-foot distance, and
$\mathrm{sign}(f)\in\{+1,-1\}$ selects the swinging leg),
and $d_{\mathrm{dcm},i} = \|\boldsymbol{\xi}_T - \mathbf{p}_i - \mathbf{b}_{\mathrm{nom}}\|^2$
is the capture-point stability residual.
These are computed from the elevation map $\mathcal{H}$
in one GPU-batched forward pass.

\paragraph{Flatness cost $Q$.}
A sole straddling a height discontinuity tilts by $\theta_i \approx \arctan(Q_i
  / L_{\mathrm{foot}})$, reducing admissible friction to $\mu_{\mathrm{eff},i}
  \leq \mu_0\cos\theta_i - \sin\theta_i$, which collapses to zero at $Q_i /
  L_{\mathrm{foot}} = \mu_0 \approx 0.6$, precisely the failure regime on
standard stairs. We use the elevation range over a footprint-sized kernel as a
GPU-efficient continuous proxy:
\begin{equation}
  Q_i = \max_{j\in\mathcal{N}_i}(z_j) - \min_{j\in\mathcal{N}_i}(z_j).
  \label{eq:flatness}
\end{equation}
Sampling-based foothold rewards~\citep{wang2025beamdojo} evaluate \emph{contact support coverage}, suited to surface voids but blind to planarity loss.
On structured terrain, a tread-edge landing scores zero under that criterion yet incurs
$Q_i > 0$; the two costs target distinct failure modes and are complementary.

\paragraph{Steepness cost $E$.}
An edge contact on a vertical riser produces a lateral impulse. Sobel-filtered
gradients $g_{x,j}$, $g_{y,j}$ are max-pooled, deliberately propagating the
worst gradient within the footprint:
\begin{equation}
  E_i = \max_{j\in\mathcal{N}_i}\!\sqrt{g_{x,j}^2 + g_{y,j}^2 + \varepsilon}.
  \label{eq:steepness}
\end{equation}

\paragraph{Height-feasibility cost $M$.}
A step too tall to reach kinematically or dynamically is worse than a
sub-optimal flat placement. The quadratic penalty
\begin{equation}
  M_i = \max\!{\left(|\Delta z_i| - h^*_{\mathrm{eff}},\;0\right)}^2,
  \label{eq:feasibility}
\end{equation}
where $\Delta z_i$ is the stance foot height difference, uses a
\emph{velocity-aware effective maximum step height}:
\begin{equation}
  h^*_{\mathrm{eff}} = h^*_{\min} + (h^*_{\max} - h^*_{\min})\,
  \clip\!\left(\dfrac{v_x}{v^*},\,0,\,1\right),
  \label{eq:heff}
\end{equation}
where $h^*_{\min}$ and $h^*_{\max}$ bound the reachable step height at standstill and
rated forward speed $v^*$, respectively.
At low speed the robot lacks the leg momentum needed
to assist a step-up, so the effective kinematic reach is reduced;
at high speed the dynamic leg drive can clear the full height.
Below a minimum speed threshold a hard override replaces the planned
target with the current foot position, and $h^*_{\mathrm{eff}} \to
  h^*_{\min}$ ensures the cost still disfavors high steps in the transition region.

\paragraph{Forward climb bonus $b$.}
To encourage the robot to step \emph{up} onto reachable treads, a speed-gated
capped bonus $b_i = \min\!\bigl(\max(\Delta z_i,
  0),\;h^*_{\mathrm{eff}}\bigr)\cdot\mathbf{1}[v_x > v_{\min}]$ rewards upward
steps proportionally to their height, capped at $h^*_{\mathrm{eff}}$.

\subsection{Bézier Swing Trajectory and Tangent-Guided Foot Orientation}%
\label{sec:swing}

\begin{figure}[t]
  \centering
  \begin{subfigure}{.30\textwidth}
    \begin{overpic}[width=1\linewidth,keepaspectratio]{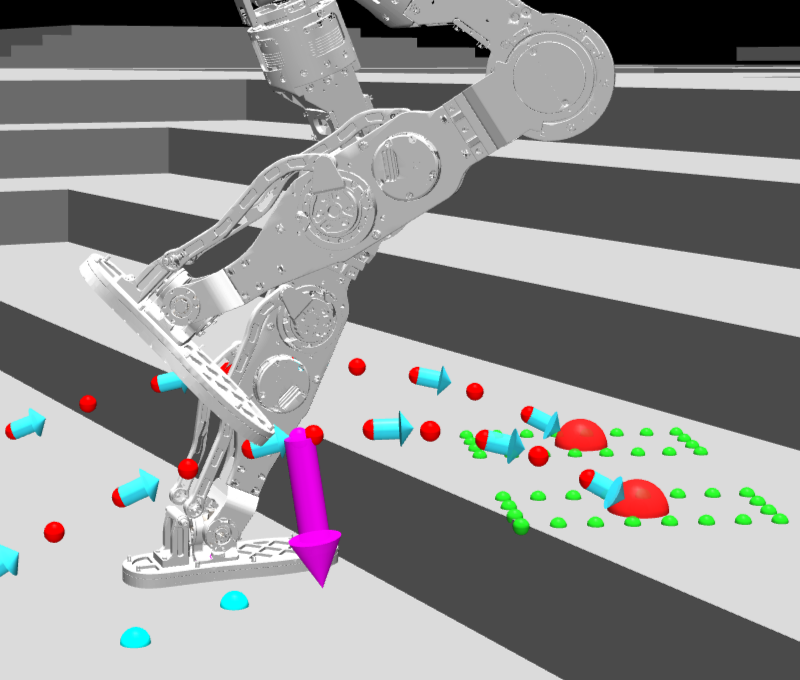}
      \put(3,3){(a)}
    \end{overpic}
  \end{subfigure}\hfill
  \begin{subfigure}{.30\textwidth}
    \begin{overpic}[width=1\linewidth,keepaspectratio]{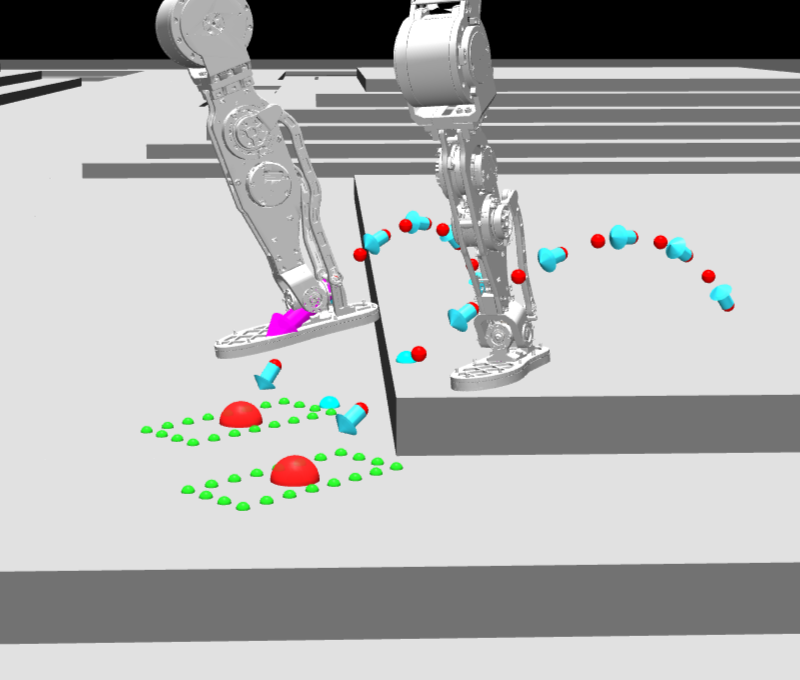}
      \put(3,3){(b)}
    \end{overpic}
  \end{subfigure}\hfill
  \begin{subfigure}{.30\textwidth}
    \begin{overpic}[width=1\linewidth,keepaspectratio]{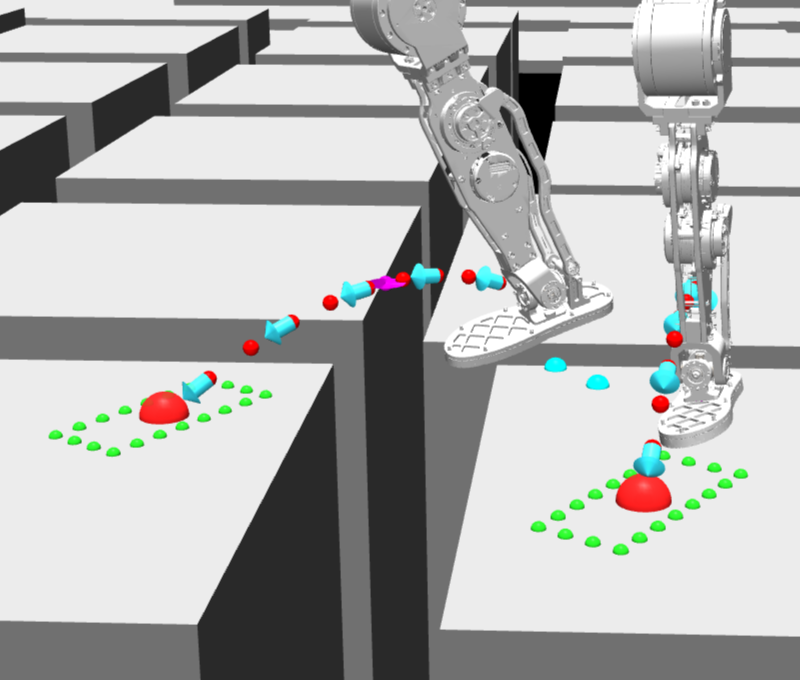}
      \put(3,3){(c)}
    \end{overpic}
  \end{subfigure}%
  \caption{\textbf{Bézier swing arcs with adaptive apex bias.}
    (a)~Step-up: apex biased toward the landing target, keeping the peak over the
    riser face.\@ (b)~Step-down: apex biased toward lift-off, extending horizontal
    travel before descent.\@ (c)~Gap: behave analogously, with clearance
    scaled by $|\Delta z|$.}%
  \label{fig:bezier}
  \vspace{-4mm}
\end{figure}

After selecting per-foot landing target $\mathbf{p}^*_f$, the swing foot tracks
a quadratic Bézier arc whose tangent simultaneously determines foot
orientation:
\begin{equation}
  \left\{\begin{aligned}
    \mathbf{p}(u)       & = {(1-u)}^2\,\mathbf{p}_\ell + 2(1-u)u\,\mathbf{p}_{\mathrm{apex}} + u^2\,\mathbf{p}^*_f,             \\
    \dot{\mathbf{p}}(u) & = 2(1-u)(\mathbf{p}_{\mathrm{apex}}-\mathbf{p}_\ell) + 2u(\mathbf{p}^*_f-\mathbf{p}_{\mathrm{apex}}),
  \end{aligned}\right.
  \quad u\in[0,1],
  \label{eq:bezier}
\end{equation}
where $\mathbf{p}_\ell$ is the lift-off position and $\dot{\mathbf{p}}(u)$ points in the
direction of instantaneous foot travel. The apex $xy$ is biased toward whichever endpoint
is higher:
\begin{equation}
  \mathrm{bias} = \clip\!\left(0.5 + \kappa\,\frac{\Delta z}{h^*_{\max}},\;b_{\min},\;b_{\max}\right),
  \quad \mathbf{p}_{\mathrm{apex},xy} = (1-\mathrm{bias})\,\mathbf{p}_{\ell,xy} + \mathrm{bias}\,\mathbf{p}^*_{f,xy},
  \label{eq:bias}
\end{equation}
For step-up ($\Delta z > 0$), bias $> 0.5$ places the trajectory peak over the riser face,
preventing premature descent onto the vertical surface; for step-down ($\Delta z < 0$),
bias $< 0.5$ extends horizontal travel before descent, reducing the landing impact angle
(\cref{fig:bezier}a-b).
A phase-conditional schedule derived from $\dot{\mathbf{p}}(u)$ guides sole orientation
through riser crossings. Both position proximity and foot heading are jointly optimized
during swing via an exponential proximity kernel:%
\label{sec:rewards}
\begin{equation}
  r_{\mathrm{foothold}} = \sum_f I_{\mathrm{swing},f}
  \exp\!\bigl(-\sigma_p\,\|\mathbf{p}_f - \mathbf{p}_{\mathrm{B\acute{e}z}}(t)\|^2
  -\sigma_d\,\|\hat{\mathbf{d}}_f - \hat{\mathbf{t}}_f(u)\|^2\bigr),
  \label{eq:rfoothold}
\end{equation}
where $\hat{\mathbf{d}}_f = R_f\hat{\mathbf{e}}_x$ is the foot forward axis and
$\sigma_d = 0$ recovers the position-only form when orientation is not used.

The apex height adapts to terrain relief:
\begin{equation}
  c = \min\!\bigl(c_{\min} + s\,|\Delta z|,\;c_{\max}\bigr),
  \quad
  z_{\mathrm{apex}} = 2\bigl(\max(z_\ell,\,z^*_f) + c\bigr)
  - \dfrac{1}{2}(z_\ell + z^*_f),
  \label{eq:clearance}
\end{equation}
where $s$ is a clearance scale factor and $c_{\max}$ caps the swing peak.
The tangent is vertical at
\begin{equation}
  u_{\mathrm{peak}} = \frac{z_\ell - z_{\mathrm{apex}}}{z_\ell - 2z_{\mathrm{apex}} + z^*_f},
  \label{eq:upeak}
\end{equation}
equal to $0.5$ for level steps and shifting toward the higher endpoint on transitions.
Two windows around $u_{\mathrm{peak}}$ define the reference unit vector
$\hat{\mathbf{t}}_f(u)$: in the \emph{pre-apex} window
$[u_{\mathrm{peak}}{-}\delta_\ell^-,\,u_{\mathrm{peak}}{-}\delta_\ell^+)$,
$\dot{\mathbf{p}}$ is rotated $\ang{90}$ in the sagittal plane
($(t_x,t_y,t_z)\!\mapsto\!(t_z,t_y,{-}t_x)$) and normalized, yielding a
forward-downward direction that guides the sole through the riser and reduces
toe-strike risk; in the \emph{post-apex} window
$(u_{\mathrm{peak}}{+}\delta_r^-,\,u_{\mathrm{peak}}{+}\delta_r^+]$,
$\hat{\mathbf{t}}_f = \dot{\mathbf{p}}/\|\dot{\mathbf{p}}\|$ aligns the foot
with the forward travel direction for tread landing; elsewhere
$\hat{\mathbf{t}}_f = \mathbf{0}$ (no orientation constraint).

\subsection{Lower Body Compliance Training}%
\label{sec:disturbance}

A payload exerts a wrench $\mathbf{W}=(\mathbf{F},\boldsymbol{\tau})$ on the
robot, where the moment
$\boldsymbol{\tau}=\mathbf{r}_\mathrm{load}\times\mathbf{F}$ scales with the
CoM-to-load arm $\mathbf{r}_\mathrm{load}$ independently of mass.
$\boldsymbol{\tau}_\mathrm{ext}$ tilts the trunk by $\varphi$, displacing the
CoM by $\Delta x_\mathrm{CoM}\approx d_\mathrm{CoM}\varphi$; under LIPM this
offset grows by $e^{\omega_0 T_\mathrm{step}}>2$ over a typical step, so
moment-induced tilt perturbs the capture point more than an equivalent
translational offset, and both components of $\mathbf{W}$ require explicit
coverage in the disturbance distribution. We emulate the wrench by applying a
spring-damper at the \emph{virtual load attachment point}
$\mathbf{p}_\mathrm{load} = \mathbf{p}_\mathrm{CoM} +
  \mathbf{r}_\mathrm{load}$:
\begin{equation}
  \mathbf{F}_\mathrm{ext}(t)
  = k\bigl(\mathbf{p}_a - \mathbf{p}_\mathrm{load}(t)\bigr)
  - c\,\dot{\mathbf{p}}_\mathrm{load}(t),
  \label{eq:msd}
\end{equation}
where $\mathbf{p}_a$ is a fixed world-frame anchor drawn once per episode, with
stiffness $k$ and damping $c$ set so the spring force at maximum displacement
$R_a$ reaches a fixed fraction of nominal weight.
Applying the force at $\mathbf{p}_\mathrm{load}$ rather than the CoM generates
$\boldsymbol{\tau}_\mathrm{ext} = \mathbf{r}_\mathrm{load}\times\mathbf{F}_\mathrm{ext}$
automatically.
At each reset, $\mathbf{r}_\mathrm{load}$ is drawn from a two-component mixture
($\mathbf{r}_\mathrm{load} = \mathbf{r}_\mathrm{body}$ with prob.\ $\rho_\mathrm{body}$,
else $\mathbf{r}_\mathrm{arm}$) that jointly covers both carry scenarios:
\emph{Body-attached} with $\mathbf{r}_\mathrm{body}\sim\mathrm{Ball}(R_\mathrm{close})$,
direction biased toward $-\hat{\mathbf{z}}$;
and \emph{Arm-extended} with $\mathbf{r}_\mathrm{arm}$ from a shoulder-frame
half-ellipsoid, biased forward.
Both components use cosine-power lobes to concentrate sampling:
\begin{equation}
  \left\{\begin{aligned}
    p(\hat{\mathbf{d}})        & \propto \cos^{n}\theta,\quad \theta\in[0,\pi/2]
                               &                                                                                                    & \text{(body-attached direction)}, \\
    p(\mathbf{r}_\mathrm{arm}) & \propto \cos^{n_r}\!\bigl(\angle(\mathbf{r}_\mathrm{arm},\hat{\mathbf{x}}_\mathrm{shoulder})\bigr)
                               &                                                                                                    & \text{(arm-extended bias)}.
  \end{aligned}\right.
  \label{eq:sampling}
\end{equation}
where $\hat{\mathbf{d}}$ is the unit direction of $\mathbf{r}_\mathrm{body}$ and
$\theta$ its polar angle from $-\hat{\mathbf{z}}$.
A weight-$\epsilon$ isotropic component blends in lateral and overhead samples.
We replace the height and orientation penalties with wrench-aware compliance
targets. For the height,
\begin{equation}
  z^*_\mathrm{virt}(t)
  = h^\mathrm{terrain}_\mathrm{pelvis}(t) + h^*
  + \alpha_z\!\left(z_a(t) - h^\mathrm{terrain}_\mathrm{pelvis}(t) - h^*\right),
  \label{eq:zvirt}
\end{equation}
where $h^*$ is the commanded base height, $z_a(t) = {[\mathbf{p}_\mathrm{load}(t)]}_z$
is the world-frame z-coordinate of the virtual attachment point, and
$\alpha_z\in[0,1]$ is the height compliance gain. For the trunk orientation, the virtual target is derived from the induced
moment:
\begin{equation}
  \varphi^*_\mathrm{virt} = \alpha_\varphi\,\frac{\tau_y}{k_\mathrm{rot}},
  \qquad
  \psi^*_\mathrm{virt}   = \alpha_\psi\,\frac{\tau_x}{k_\mathrm{rot}},
  \label{eq:tilt_target}
\end{equation}
where $\tau_x,\tau_y$ are the roll and pitch components of $\boldsymbol{\tau}_\mathrm{ext}$,
$k_\mathrm{rot}$ $[\mathrm{N{\cdot}m/rad}]$ is a rotational stiffness converting induced moment to
a tilt angle, and $\alpha_\varphi,\alpha_\psi\in[0,1]$ are orientation compliance gains. The combined compliance reward is
\begin{equation}
  r_\mathrm{comply}
  = -{\bigl(z_\mathrm{pelvis} - z^*_\mathrm{virt}\bigr)}^2
  - {\bigl(\varphi_B - \varphi^*_\mathrm{virt}\bigr)}^2
  - {\bigl(\psi_B   - \psi^*_\mathrm{virt}\bigr)}^2,
  \label{eq:rcomply}
\end{equation}
where $\varphi_B,\psi_B$ are pelvis pitch and roll from the projected gravity
vector. When $\alpha_z = \alpha_\varphi = \alpha_\psi = 0$ the reward collapses to the
original rigid height and orientation penalties; at nonzero gains the policy learns to yield to the full
wrench, acquiring wrench-aware compliance in both translation and rotation as a
learned behavior without explicit force or torque sensing.

\subsection{Policy and Training}%
\label{sec:policy}
An asymmetric Actor-Critic Encoder is used. The actor concatenates a
convolutional embedding of a temporally stacked depth image with proprioceptive
state and passes the result through a feedforward multilayer perceptron (MLP)\@. The critic shares the
same encoder but also receives privileged training-only signals.
Actions are joint position targets relative to the default pose, converted to
torques via proportional-derivative (PD) control and clipped to mechanical limits. Following the
gait-adaptive extension~\citep{song2025gaitadaptive}, the policy also outputs a
scalar gait frequency $f_t$ that advances an internal phase clock via an
exponential-moving-average (EMA) smoothed update:
\begin{equation}
  \hat{f}_t = (1-\alpha_{\mathrm{EMA}})\hat{f}_{t-1} + \alpha_{\mathrm{EMA}}\,\clip(f_t,\,f_{\min},\,f_{\max}),
  \quad
  \phi_{t+1} = (\phi_t + \Delta t\,\hat{f}_t)\bmod 1,
  \label{eq:gaitfreq}
\end{equation}
with $(\hat{f}_t,\sin 2\pi\phi_t,\cos 2\pi\phi_t)$ appended to the actor
observation. Training uses standard PPO~\citep{schulman2017ppo} with adaptive
Kullback--Leibler (KL) scheduling, generalized advantage estimation (GAE), and
massively-parallel joint-velocity and terrain-difficulty
curricula~\citep{pmlr-v164-rudin22a}.

\section{Results}%
\label{sec:results}

\paragraph{Experimental setup.}
Four variants are compared: \emph{TACT\,+\,Adaptive Gait (Ours)},
\emph{TACT-only}, \emph{Adaptive Gait only}, and \emph{Baseline} (a standard
depth-map perceptive policy with no terrain-cost channels and no privileged
elevation-map input to the critic). Each variant is evaluated at iteration
20\,k across 4096 environments in MuJoCo~\citep{todorov2012mujoco} on stairs,
slopes, and rough terrain; a second, unseen hard-terrain sweep covers risers
$0.20$--$0.30$\,m, over which we report the success rate
$\mathrm{SR}_\mathrm{hard}$ computed on the strict interior $[0.22,0.28]$\,m.

\subsection{Terrain Traversal Ablation}%
\label{sec:traversal}

\begin{figure}[t]
  \centering
  \includegraphics[width=\textwidth]{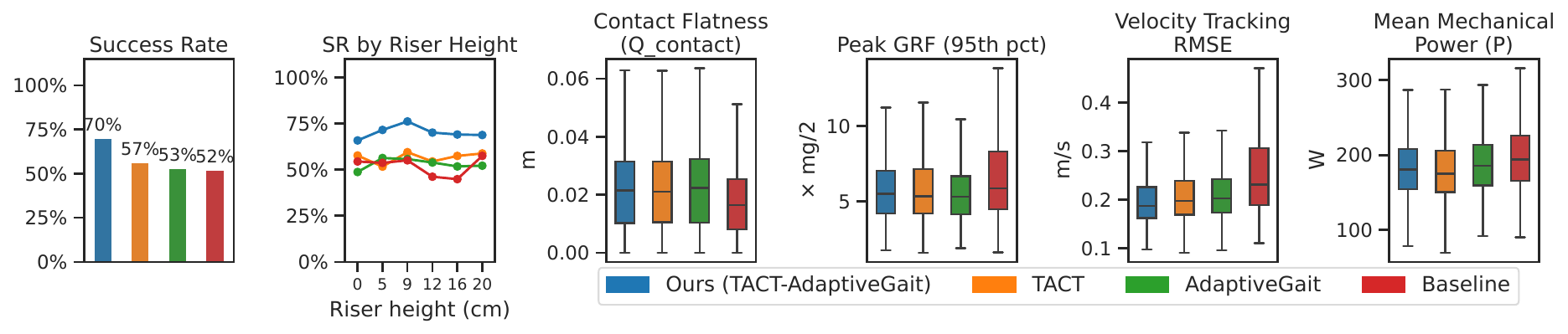}\\
  \includegraphics[width=\textwidth]{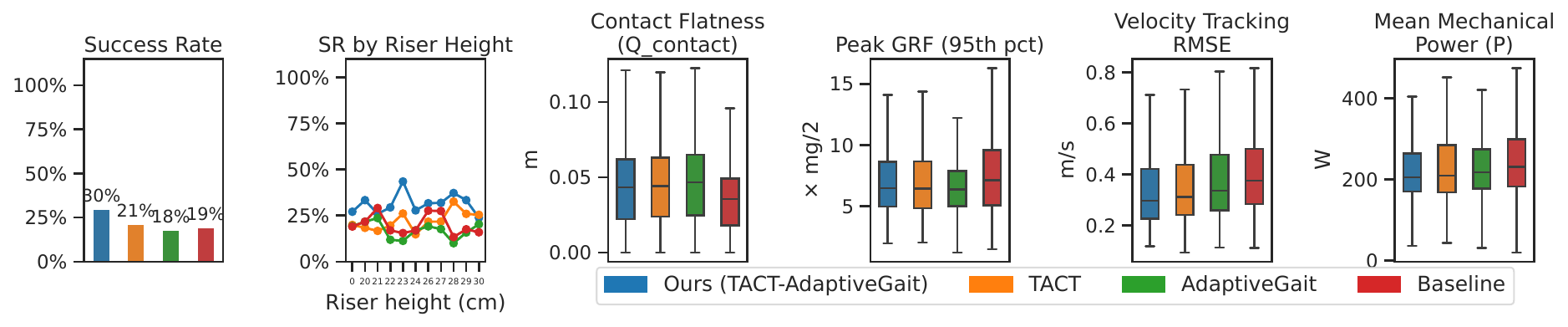}
  \caption{Terrain traversal ablation on \textbf{standard terrain} (top) and \textbf{hard terrain} (out-of-distribution).}%
  \label{fig:ablation}
  \vspace{-3mm}
\end{figure}

\begin{figure}[t]
  \centering
  \begin{subfigure}[t]{.32\textwidth}
    \centering
    \begin{overpic}[height=3cm,keepaspectratio]{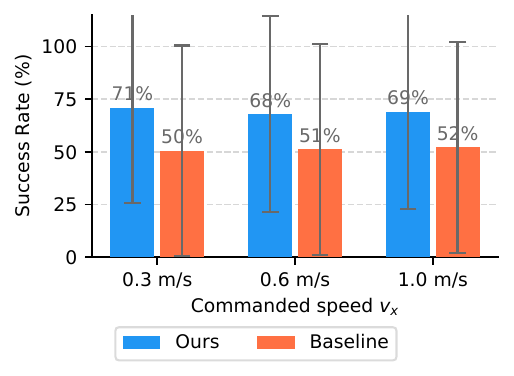}
      \put(3,5){(a)}
    \end{overpic}%
    \phantomcaption{}\label{fig:speed_sr}
  \end{subfigure}\hfill
  \begin{subfigure}[t]{.32\textwidth}
    \centering
    \begin{overpic}[height=3cm,keepaspectratio]{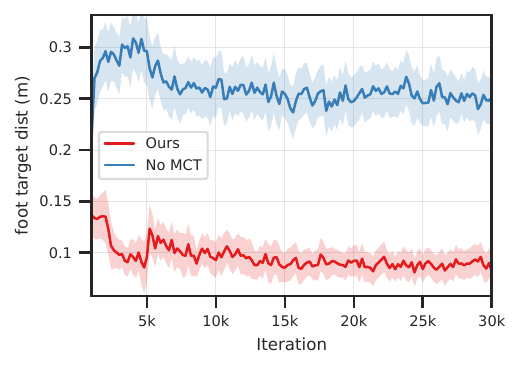}
      \put(3,5){(b)}
    \end{overpic}%
    \phantomcaption{}\label{fig:foot_target_dist_ablation}
  \end{subfigure}\hfill
  \begin{subfigure}[t]{.32\textwidth}
    \centering
    \begin{overpic}[height=3cm,keepaspectratio]{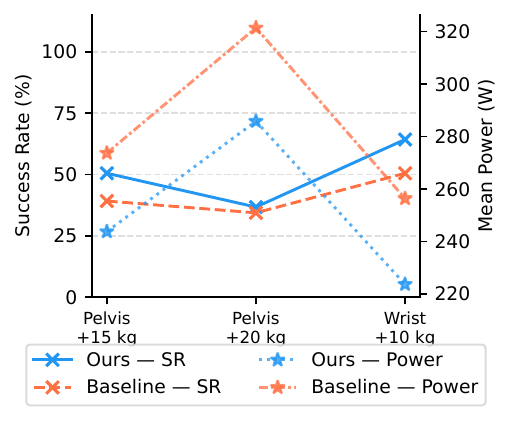}
      \put(-5,5){(c)}
    \end{overpic}%
    \phantomcaption{}\label{fig:payload_sr_power}
  \end{subfigure}\\%
  \begin{subfigure}{.32\textwidth}
    \centering
    \begin{overpic}[height=3cm,keepaspectratio]{figures/F6.Experiment_down.png}
      \put(-15,5){(d)}
    \end{overpic}
    \phantomcaption{}\label{fig:real_exp_up}
  \end{subfigure}\hfill
  \begin{subfigure}{.32\textwidth}
    \centering
    \begin{overpic}[height=3cm,keepaspectratio]{figures/F6.Experiment_up.png}
      \put(-15,5){(e)}
    \end{overpic}
    \phantomcaption{}\label{fig:real_exp_down}
  \end{subfigure}\hfill
  \begin{subfigure}{.32\textwidth}
    \centering
    \begin{overpic}[height=3cm,keepaspectratio]{figures/F1.Teaser_2.png}
      \put(-12,5){(f)}
    \end{overpic}
    \phantomcaption{}\label{fig:real_exp_wrist}
  \end{subfigure}
  \caption{(a)~Speed-conditioned SR;\@ (b)~foot-target
    distance; (c)~SR (\%) and mean power (W).\@ (d--f)~Qualitative hardware payload
    demonstrations, separate from the simulated ablations (d--e: 20\,kg backpack; f: 7\,kg tray).
  }%
  \label{fig:speed_ablation}
  \vspace{-3mm}
\end{figure}
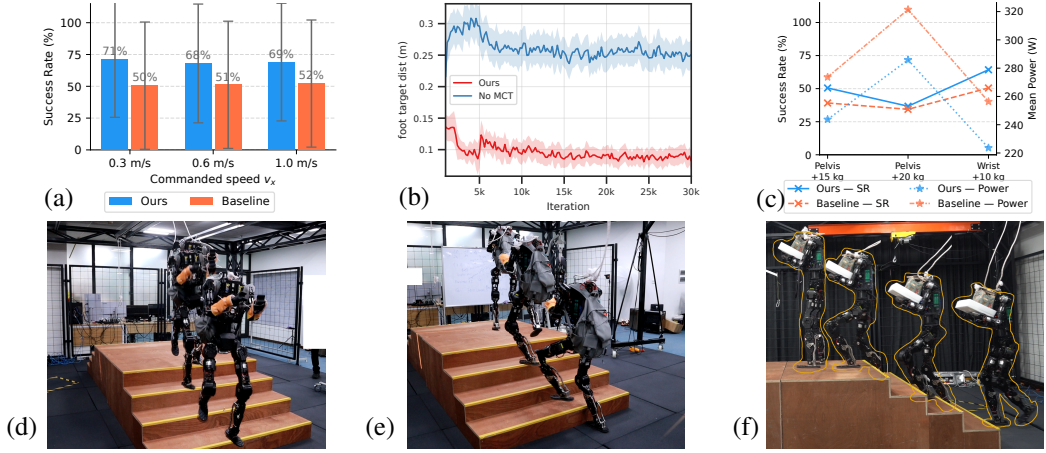

Ours (TACT\,+\,Adaptive Gait) leads on both standard and hard terrain (70\,\%
standard SR and 30\,\% $\mathrm{SR}_\mathrm{hard}$), outperforming TACT-only by
13 and 9\,percentage points~(pp) and Adaptive Gait only by 17 and 12\,pp
(Fig.~\ref{fig:ablation}); Adaptive Gait only falls below the Baseline on hard
terrain (18\,\% vs.\ 19\,\%), confirming that frequency re-timing without
terrain-quality guidance is actively harmful when risers approach kinematic
limits. The Baseline's 52\,\% and 19\,\% SR coincide with the highest
95th-percentile ground reaction force (GRF) (${\approx}6.7\!\times$ and
$8.0\!\times\!mg/2$), consistent with edge-biased contacts from blind foothold
placement. On efficiency, Ours achieves lower velocity-tracking
root-mean-square error (RMSE) ($0.22$ vs.\ $0.23\,\mathrm{m/s}$) and mechanical
power ($229$ vs.\ $241\,\mathrm{W}$) than TACT-only on hard terrain;
$Q_\mathrm{c}$ is identical between the two ($0.023$), isolating terrain cost
channels as the mechanism for tread-centered landings. Fig.~\ref{fig:speed_sr}
shows that the SR gap relative to the Baseline is speed-invariant
($71\,/\,68\,/\,69\,\%$ vs.\ ${\approx}20\,\mathrm{pp}$ lower at
$0.3\,/\,0.6\,/\,1.0\,\mathrm{m/s}$), confirming foothold quality, not speed,
drives the difference; Fig.~\ref{fig:foot_target_dist_ablation} shows that
removing all TACT channel weights raises foot-target distance $2.8\!\times$
($0.088\!\to\!0.251\,\mathrm{m}$).

\subsection{Payload Generalization}%
\label{sec:payload_results}

Fig.~\ref{fig:payload_sr_power} evaluates payload generalization across three
conditions (pelvis $+$15\,kg, pelvis $+$20\,kg, wrist $+$10\,kg) on terrain and
compares Ours against the Baseline. At moderate centered load (pelvis
$+$15\,kg), Ours achieves 50\,\% SR versus the Baseline's 38\,\% at lower power
(247\,W vs.\ 277\,W), consistent with compliance training absorbing the
downward wrench and suppressing impulsive GRF recovery. At high centered load
(pelvis $+$20\,kg), both policies degrade substantially to $\approx$37\,\% and
$\approx$35\,\% SR respectively, nearly indistinguishable, indicating the load
magnitude approaches the boundary of the training distribution; power rises for
both, with Ours remaining 27\,W below the Baseline (293 vs.\ 320\,W),
confirming that partial compliance is retained even as balance recovery
deteriorates. Wrist-mounted mass ($+$10\,kg), which generates a large moment
rather than a direct CoM shift, yields the largest margin across all three
conditions: Ours achieves 65\,\% SR versus the Baseline's 50\,\% at 223\,W
versus 259\,W, consistent with the arm-extended wrench samples in the
disturbance force-field explicitly targeting moment-dominated loads.

\section{Limitations}%
\label{sec:limitations}

\textbf{Simulation-only evaluation.} All quantitative ablations are conducted in
MuJoCo; hardware results are qualitative demonstrations, with domain
randomization as the sole sim-to-real bridge, so absolute success rates on real
structured terrain remain uncharacterized and unmodeled contact phenomena will
shift them. Even in simulation, the best variant reaches only 30\,\%
$\mathrm{SR}_\mathrm{hard}$ on hard terrain (interior risers $0.22$--$0.28$\,m),
a ceiling insufficient for reliable deployment on tall staircases.

\textbf{Compliance and perception scope.} The compliance benefit vanishes near
20\,kg of centered load (Ours $\approx$37\,\% vs.\ Baseline $\approx$35\,\% SR at
pelvis $+$20\,kg): the sampled wrench distribution and leg-only actuation set an
effective payload ceiling, and lateral or distributed loads fall outside the
sampled space. The DCM planner depends on accurate elevation-map registration
with no fallback under localization drift, and the forward-facing depth camera
gives no lateral coverage. Finally, the terrain cost channels are collectively
necessary but individually non-critical, removing any one keeps foot-target
distance within 5\,\%, and their weights are hand-tuned rather than derived.
Extended analysis of each limitation (adaptive-gait fragility, channel
redundancy, upper-body coupling, and deformable/low-friction terrain) is given in
Appendix.

\section{Conclusion}%
\label{sec:conclusion}

We presented a physics-informed terrain affordance learning system for
payload-robust perceptive humanoid locomotion. A multi-channel terrain cost
drives a GPU-parallel DCM foothold planner and provides a dense
terrain-affordance learning signal to the RL policy; a terrain-adaptive Bézier
swing with tangent-guided foot orientation extends foothold tracking to joint
position-and-orientation references. Because the cost is defined from sole
geometry rather than learned per embodiment~\citep{miki2022wild}, it is
platform-agnostic by construction. Lower-body compliance training via a
cosine-power-weighted virtual spring-damper yields a learned wrench-dependent
impedance (conceptually related to virtual impedance
tracking~\citep{xu2025facetforceadaptivecontrolimpedance} but without a
separate impedance reference, and avoiding the online payload estimation
of~\citet{kumar2021rma}), improving payload robustness up to
${\sim}15\,\mathrm{kg}$ centered load and for moment-dominated wrist loads
without retraining. The full system trains with standard PPO in a single run
(no teacher-student
distillation~\citep{agarwal2022leggedlocomotionchallengingterrains,zhang2026rpl})
and is deployed zero-shot from simulation on a humanoid; quantitative
real-world evaluation and a terrain$\,\times\,$payload coupling study are the
main remaining steps toward reliable deployment.

\clearpage
\bibliography{references}

\appendix

\clearpage
\section*{Appendix}%
\label{app:appendix}
\renewcommand{\thesubsection}{\Alph{subsection}}
\setcounter{subsection}{0}

\subsection{Implementation Details}%
\label{app:impl}

\paragraph{DCM derivation (\cref{sec:dcm}).}\label{app:dcm}
\citet{liu2026faststair} show that for a linear CoM height profile
$z(t) = k_z t + z_0$ during swing, the Variable-Height Inverted Pendulum dynamics
reduce to $\dot{\xi} \approx \omega(t)\,\xi$ when
$a = 1 + k_z/(2\sqrt{gz_0}) \approx 1$; we adopt this result and set
$\omega=\omega_0$. The steady-state nominal step offset that keeps the divergent
mode bounded~\citep{koolen2012capturability} is $b_x = v_x T/(e^{\omega_0 T}-1)$
and $b_y = \mathrm{sign}(f)\,l_p/(1 + e^{\omega_0 T})$, where $l_p$ is the nominal
lateral inter-foot distance and $\mathrm{sign}(f)\in\{+1,-1\}$ selects the
swinging leg.

\begin{table}[h!]
  \centering
  \caption{DCM foothold planner configuration.}
  \small
  \begin{tabular}{p{0.20\linewidth}p{0.22\linewidth}p{0.46\linewidth}}
    \toprule
    Parameter                                                  & Value                            & Notes                                                                               \\
    \midrule
    $z_0$                                                      & $0.90\,\mathrm{m}$               & Nominal CoM height                                                                  \\
    $T_{\mathrm{swing}}$ (= $T$ in \cref{sec:dcm})             & $0.45\,\mathrm{s}$               & Half gait period                                                                    \\
    $l_p$                                                      & $0.20\,\mathrm{m}$               & Lateral inter-foot distance                                                         \\
    $\kappa$                                                   & $0.4$                            & Bézier apex bias gain                                                               \\
    $b_{\min},\,b_{\max}$                                      & $0.25,\;0.75$                    & Bézier bias clamp bounds                                                            \\
    $c_{\min}$                                                 & $0.05\,\mathrm{m}$               & Base swing clearance                                                                \\
    $s$                                                        & $0.5$                            & Clearance scale (m per m step height)                                               \\
    $c_{\max}$                                                 & $0.20\,\mathrm{m}$               & Maximum swing clearance                                                             \\
    $\delta_\ell^-, \delta_\ell^+$                             & $0.30,\;0.05$                    & Lift phase: swing fractions before/after apex where tangent orientation is blended  \\
    $\delta_r^-, \delta_r^+$                                   & $0.05,\;0.25$                    & Reach phase: swing fractions before/after apex where landing orientation is blended \\
    Elevation map                                              & $37\times25$, $0.05\,\mathrm{m}$ & $1.80\times1.20\,\mathrm{m}$ at pelvis                                              \\
    Foot footprint kernel                                      & $0.25\times0.10\,\mathrm{m}$     & For $Q_i$, $E_i$ pooling                                                            \\
    Search window                                              & $\pm0.30\,\mathrm{m}$ (x and y)  & Around nominal stride; fallback to nominal at mean valid height if empty            \\
    $h^*_{\min}$                                               & $0.05\,\mathrm{m}$               & Effective step height at standstill                                                 \\
    $h^*_{\max}$                                               & $0.28\,\mathrm{m}$               & Effective step height at rated speed                                                \\
    $v^*$                                                      & $0.5\,\mathrm{m/s}$              & Rated forward speed for $h^*_{\mathrm{eff}}$ scaling (Eq.~\eqref{eq:heff})          \\
    $v_{\min}$                                                 & $0.05\,\mathrm{m/s}$             & Standing override: planned target replaced by current foot position                 \\
    Lateral penalty                                            & $\beta = 2.5$                    & Lateral vs.\ sagittal asymmetry                                                     \\
    \midrule
    $\alpha_{\mathrm{pos}},\,\alpha_{\mathrm{dcm}}$            & $1.0,\;0.5$                      & Position and capture-point stability                                                \\
    $\alpha_E,\,\alpha_Q,\,\alpha_M,\,\alpha_{\mathrm{climb}}$ & $0.6,\;4.0,\;6.0,\;1.5$          & Terrain channels (Eq.~\eqref{eq:cost})                                              \\
    \midrule
    \multicolumn{3}{l}{\textit{Gait-adaptive extension}}                                                                                                                                \\
    $f_{\min},\,f_{\max}$                                      & $1.0,\;1.5\,\mathrm{Hz}$         & Gait frequency bounds                                                               \\
    $\alpha_{\mathrm{EMA}}$                                    & $0.2$                            & Frequency smoothing weight                                                          \\
    \bottomrule
  \end{tabular}
\end{table}

\begin{table}[h!]
  \centering
  \caption{Lower-body compliance training parameters (\cref{sec:disturbance}).}
  \small
  \begin{tabular}{lll}
    \toprule
    Parameter                                             & Value                              & Notes                                                                   \\
    \midrule
    \multicolumn{3}{l}{\textit{Spring-damper}}                                                                                                                           \\
    $R_a$                                                 & $0.40\,\mathrm{m}$                 & Anchor ball radius                                                      \\
    $k$                                                   & $200\,\mathrm{N/m}$                & Virtual spring stiffness                                                \\
    $c$                                                   & $20\,\mathrm{N{\cdot}s/m}$         & Virtual damping                                                         \\
    \midrule
    \multicolumn{3}{l}{\textit{Wrench sampling}}                                                                                                                         \\
    $\rho_\mathrm{body}$                                  & $0.6$                              & Body-attached component probability                                     \\
    $R_\mathrm{close}$                                    & $0.10\,\mathrm{m}$                 & Body-attached ball radius                                               \\
    $n$                                                   & $2$                                & Cosine-power exponent (bias toward $-\hat{\mathbf{z}}$)                 \\
    $n_r$                                                 & $3$                                & Cosine-power exponent (arm-extended forward bias)                       \\
    $(a_\mathrm{fwd},\,a_\mathrm{lat},\,a_\mathrm{vert})$ & $(0.50,\,0.40,\,0.30)\,\mathrm{m}$ & Arm-extended half-ellipsoid semi-axes                                   \\
    $\epsilon$                                            & $0.1$                              & Isotropic mixture weight                                                \\
    \midrule
    \multicolumn{3}{l}{\textit{Compliance targets (Eq.~\eqref{eq:zvirt},~\eqref{eq:tilt_target})}}                                                                       \\
    $k_\mathrm{leg}$                                      & $300\,\mathrm{N/m}$                & Effective leg stiffness; gives $\alpha_z = k/(k+k_\mathrm{leg}) = 0.40$ \\
    $\alpha_z$                                            & $k/(k+k_\mathrm{leg}) = 0.40$      & Height compliance gain                                                  \\
    $k_\mathrm{rot}$                                      & $50\,\mathrm{N{\cdot}m/rad}$       & Rotational stiffness (moment-to-tilt conversion)                        \\
    $\alpha_\varphi,\,\alpha_\psi$                        & $0.5$                              & Orientation compliance gains (pitch, roll)                              \\
    \bottomrule
  \end{tabular}
\end{table}

\begin{table}[h!]
  \centering
  \caption{Reward terms. Terrain-specific terms are highlighted; all other terms are standard
    locomotion rewards. $I_f$ denotes foot contact, $F_f$ contact force, $L_c$ centroidal
    angular momentum.}%
  \label{tab:rewards}
  \small
  \begin{tabular}{p{0.20\linewidth}rp{0.50\linewidth}}
    \toprule
    Term                     & $w$             & Definition                                                                                               \\
    \midrule
    \multicolumn{3}{l}{\textit{Velocity tracking}}                                                                                                        \\
    $r_{\mathrm{vel},xy}$    & $+3.5$          & $\exp(-\|e_{xy}\|^2/0.25)$                                                                               \\
    $r_{\mathrm{vel},z}$     & $+3.0$          & $\exp(-e_{z}^2/0.5)$                                                                                     \\
    \midrule
    \multicolumn{3}{l}{\textit{Gait quality}}                                                                                                             \\
    $r_{\mathrm{gait}}$      & $+2.0$          & Phase match: stance/swing vs.\ gait clock                                                                \\
    $r_{\mathrm{airtime}}$   & $+2.0$          & Single-stance duration reward                                                                            \\
    $r_{\mathrm{orient}}$    & $-7.0$          & $\|\mathbf{g}_{xy}\|^2$ (base tilt)                                                                      \\
    $r_{\mathrm{pelvis}}$    & $-3.0$          & $\|\mathbf{g}_{xy}\|^2$ (pelvis)                                                                         \\
    $r_{\mathrm{action}}$    & $-0.8$          & $\|a_t - a_{t-1}\|^2$                                                                                    \\
    $r_{\mathrm{collision}}$ & $-2.0$          & Self-collision force $> 10\,\mathrm{N}$                                                                  \\
    $r_{\mathrm{limits}}$    & $-1.0$          & Joint-limit violation                                                                                    \\
    $r_{\mathrm{slip}}$      & $-0.4$          & Tangential contact velocity                                                                              \\
    $r_{L}$                  & $-0.001$        & $\|L_c\|^2$ angular momentum                                                                             \\
    \midrule
    \multicolumn{3}{l}{\textit{\textbf{Terrain-specific (contribution)}}}                                                                                 \\
    $r_{\mathrm{foothold}}$  & $\mathbf{+2.1}$ & Eq.~\eqref{eq:rfoothold};\; $\sigma_p=10$,\; $\sigma_d\in\{0,\,5.0\}$                                    \\
    $r_{\mathrm{clearance}}$ & $\mathbf{-0.5}$ & $\sum_f |h_f^{\mathrm{foot}} - h_f^{\mathrm{terrain}} - d^*|\cdot\|v_{f,xy}\|$,\; $d^*=0.06\,\mathrm{m}$ \\
    $r_{\mathrm{stumble}}$   & $\mathbf{-0.5}$ & $\mathbf{1}[\|F^{xy}_f\|>4|F^z_f|\;\wedge\;\|v^{xy}_f\|>0.15\,\mathrm{m/s}]$                             \\
    $r_{\mathrm{comply}}$    & $\mathbf{-1.5}$ & Eq.~\eqref{eq:rcomply};\; $h^*=0.90\,\mathrm{m}$,\; $\alpha_z = k/(k+k_\mathrm{leg})$                    \\
    \bottomrule
  \end{tabular}
\end{table}

\begin{table}[h!]
  \centering
  \caption{Domain randomization ranges applied at episode startup.}%
  \label{tab:dr}
  \small
  \begin{tabular}{lll}
    \toprule
    Parameter                 & Range                                                                                                & Mode     \\
    \midrule
    Waist-link mass offset    & $[-2, +2]\,\mathrm{kg}$                                                                              & add      \\
    Other body masses         & $\times[0.95, 1.05]$                                                                                 & scale    \\
    Foot friction             & $[0.3, 1.6]$                                                                                         & absolute \\
    PD gains ($K_p$, $K_d$)   & $\times[0.7, 1.1]$                                                                                   & scale    \\
    Joint stiffness / damping & $\times[0.7, 1.3]$                                                                                   & scale    \\
    Joint armature            & $\times[0.2, 5.0]$                                                                                   & scale    \\
    CoM offset (pelvis)       & $\pm 0.05\,\mathrm{m}$                                                                               & add      \\
    Encoder bias              & $\pm 0.015\,\mathrm{rad}$ per joint                                                                  & add      \\
    Depth camera pose         & pos $x,y\;\pm0.01$\,m, $z\;({-}0.03,{+}0.01)$\,m; rot roll/yaw $\pm2^{\circ}$, pitch $\pm10^{\circ}$ & add      \\
    \bottomrule
  \end{tabular}
\end{table}

\subsection{Policy Learning Details}%
\label{app:policy}

\paragraph{Observation space (actor).}
Base angular velocity, projected gravity vector, velocity command, joint
positions and velocities relative to default, last action, binary foot
contacts, gait phase ($\sin\phi$, $\cos\phi$), and a stacked depth image (four
temporally delayed frames at $30\,\mathrm{Hz}$, blurred, range-clipped, and
normalized). Proprioceptive signals are maintained as a 5-frame history at the
$50\,\mathrm{Hz}$ control frequency; the depth image stack is updated at the
camera pipeline rate ($30\,\mathrm{Hz}$).

\paragraph{Observation space (critic, training only).}
Actor observations plus: true CoM velocity, noiseless elevation map (ray misses
filled with a sentinel value), ground-truth foot contact forces and heights,
foot air times, and planner landing targets $\mathbf{p}^*_f$ for each foot.

\begin{table}[h!]
  \centering
  \caption{Actor network specification.}
  \small
  \begin{tabular}{ll}
    \toprule
    Component     & Specification                                                                             \\
    \midrule
    Depth encoder & Conv2d: channels $[32,32]$, kernel $3\times3$, max-pool                                   \\
                  & FC:\@ $[128, 64]$, output embedding $\in\R^{64}$                                          \\
    Actor MLP     & Input: embed (64) $\oplus$ proprioception; hidden $(1024, 512, 256, 128)$, ELU activation \\
    Action        & Joint position increments, clipped to limits                                              \\
    History       & 5-frame obs history at 50\,Hz                                                             \\
    \bottomrule
  \end{tabular}
\end{table}

\paragraph{Training hyperparameters.}
PPO with clip parameter $\epsilon = 0.2$, learning rate $10^{-3}$ with adaptive
KL-based scheduling, GAE $\lambda = 0.95$, discount $\gamma = 0.99$, 5 learning
epochs per rollout, 4 mini-batches, 24 steps per environment per rollout.

\paragraph{Curriculum.}
A two-stage velocity curriculum starts at $[0, 0.5]\,\mathrm{m/s}$ and expands
to $[0, 1.0]\,\mathrm{m/s}$ after 120{,}000 steps. The terrain curriculum uses
stairs with 0.05--0.20\,m risers and 0.30--0.60\,m treads, open-width stairs,
and flat sections.

\paragraph{Gait-adaptive frequency action.}
The policy outputs a scalar gait frequency $f_t \in [f_{\min}, f_{\max}]$ as an
additional action dimension appended after joint targets. Each physics step the
gait phase advances as $\phi_t = (\phi_{t-1} + \Delta t_{\mathrm{phys}} \cdot
  f_t) \bmod 1.0$. The raw policy output is clipped to $[f_{\min}, f_{\max}]$ and
smoothed by an EMA before use: $f_t = (1{-}\alpha_{\mathrm{EMA}})\,f_{t-1} +
  \alpha_{\mathrm{EMA}}\,\clip(f_t^{\mathrm{raw}}, f_{\min}, f_{\max})$. The
phase and its sinusoidal encoding $(\sin\phi_t,\cos\phi_t)$ are included in the
actor observation; $f_{\min}$, $f_{\max}$, and $\alpha_{\mathrm{EMA}}$ are
listed in the DCM planner table (\cref{app:impl}).

\paragraph{Auxiliary reward terms.}\label{app:aux_rewards}
\textit{Clearance}: $\sum_f |h_f^{\mathrm{foot}} - h_f^{\mathrm{terrain}} -
  d^*|\cdot\|v_{f,xy}\|$ penalizes swing proximity to the terrain,
velocity-weighted so edge collisions at high swing speed incur proportionally
larger cost.
\textit{Stumble}: fires when $\|F^{xy}_f\| > 4|F^z_f|$ and $\|v^{xy}_f\| >
  0.15\,\mathrm{m/s}$; the criterion $\mu_{\mathrm{eff}} > 4$ reliably
distinguishes riser edge-strikes from high-force landings.
\textit{Compliant base height}: ${(z_{\mathrm{pelvis}} - z^*_{\mathrm{virt}})}^2$
tracks a virtual compliant target (\cref{eq:zvirt}); at zero compliance gain it
reduces to the standard terrain-relative penalty, preserving the LIPM
eigenfrequency $\omega_0 = \sqrt{g/z_0}$ through elevation changes.

\subsection{Experimental Details}%
\label{app:experiments}

\paragraph{Training setup.}
All policies are trained on a single NVIDIA H100 SXM5 80\,GB GPU with 8192
parallel MuJoCo environments. Each iteration collects 24 simulation steps per
environment. The PPO rollout and network update run on the same GPU;\@ physics
simulation is CPU-parallelized across the host node's available cores. All
training uses the \texttt{mjlab} framework with MuJoCo 3.8.0 as the physics
backend. Total wall-clock training time per variant is approximately 10--12\,h.

\begin{figure}[h!]
  \centering
  \begin{subfigure}{.23\textwidth}
    \begin{overpic}[width=1\linewidth,keepaspectratio]{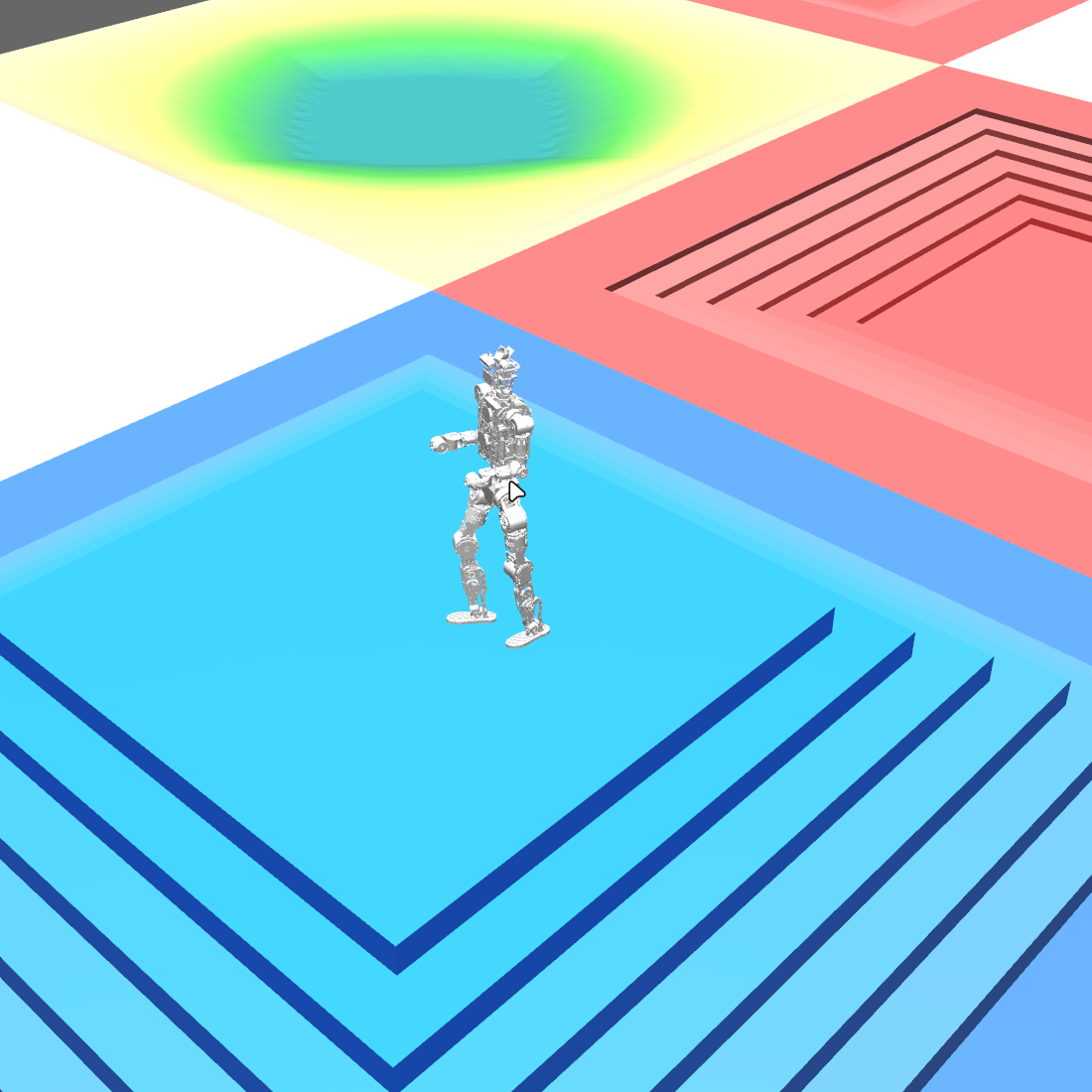}
      \put(3,3){(a)}
    \end{overpic}
  \end{subfigure}\hfill
  \begin{subfigure}{.23\textwidth}
    \begin{overpic}[width=1\linewidth,keepaspectratio]{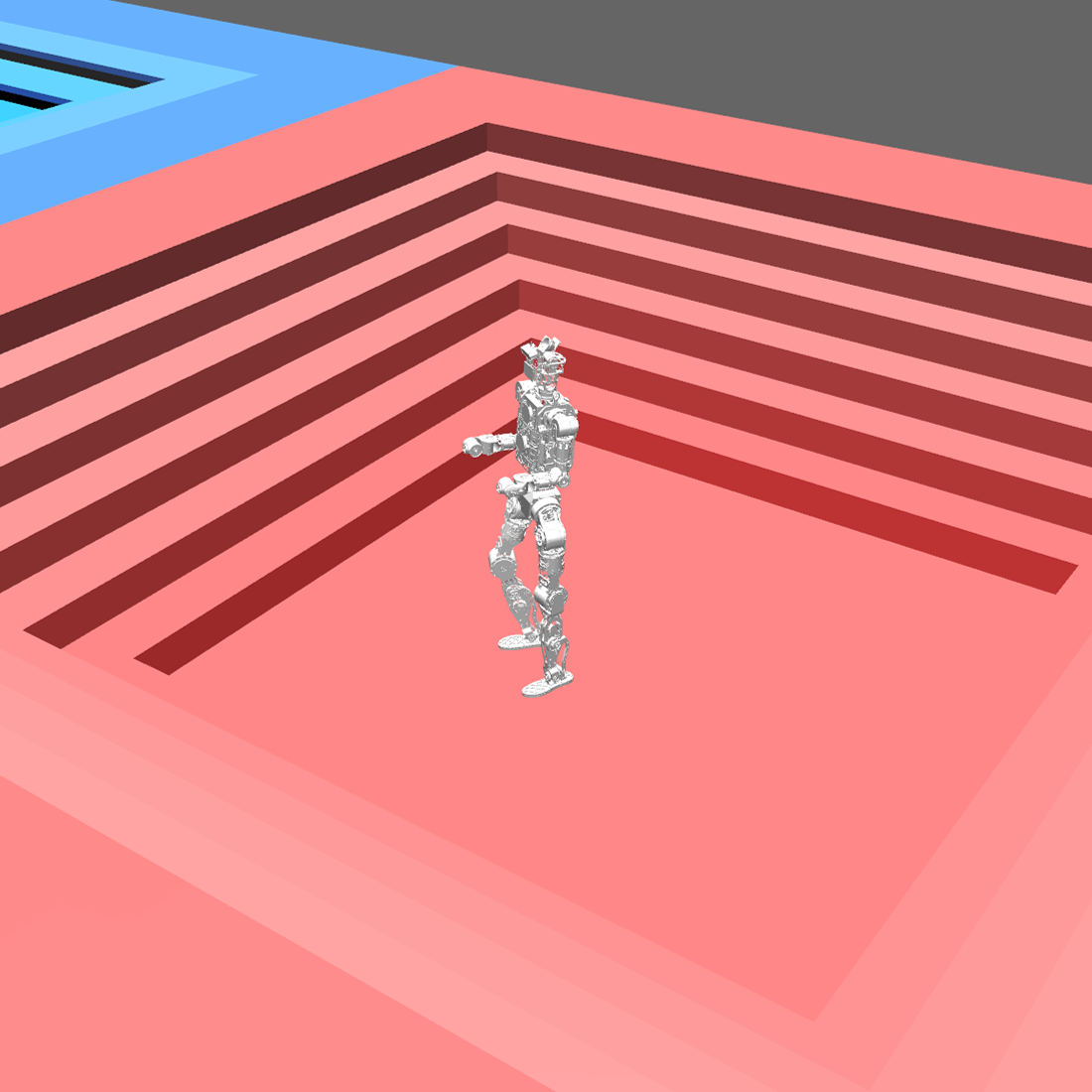}
      \put(3,3){(b)}
    \end{overpic}
  \end{subfigure}\hfill
  \begin{subfigure}{.23\textwidth}
    \begin{overpic}[width=1\linewidth,keepaspectratio]{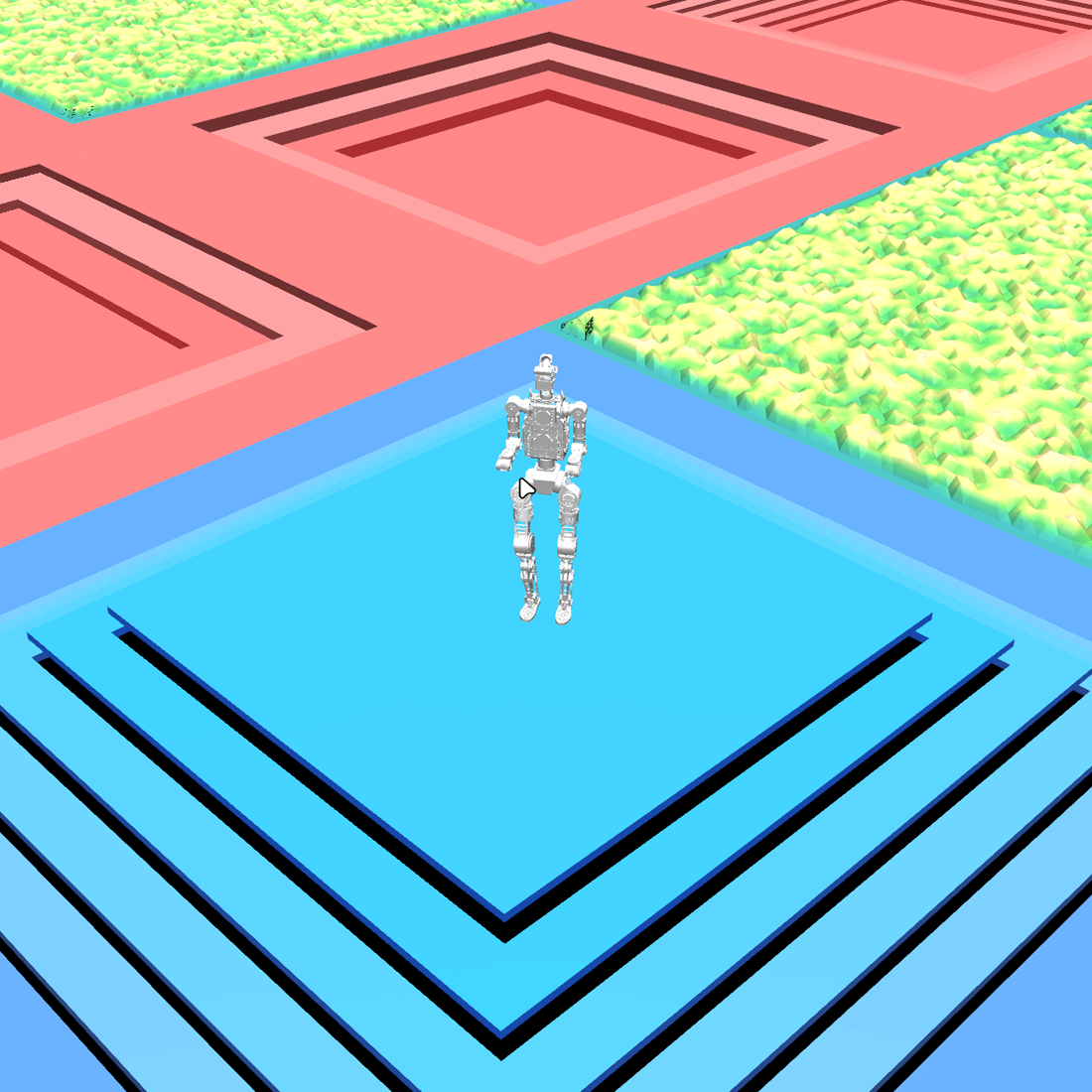}
      \put(3,3){(c)}
    \end{overpic}
  \end{subfigure}\hfill
  \begin{subfigure}{.23\textwidth}
    \begin{overpic}[width=1\linewidth,keepaspectratio]{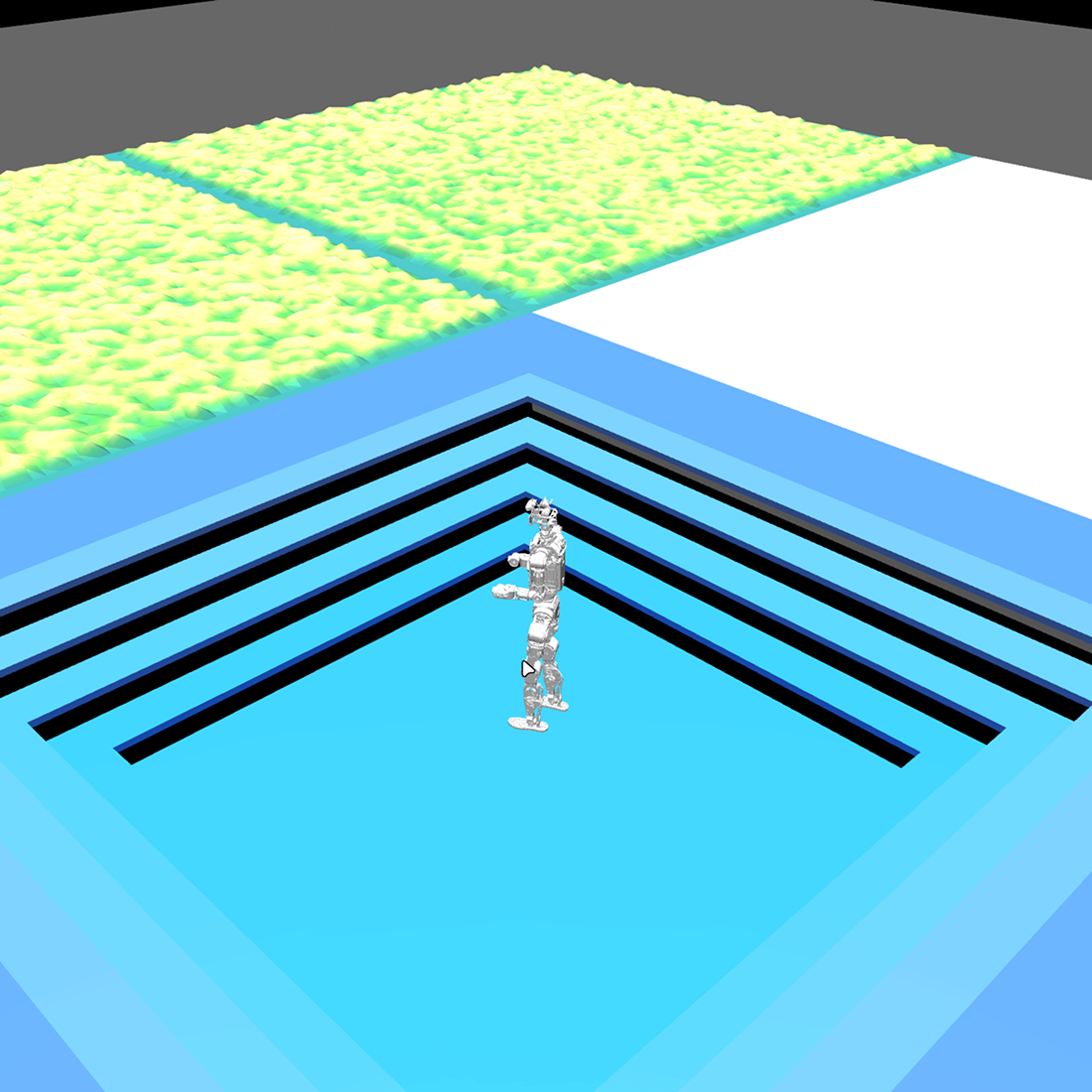}
      \put(3,3){(d)}
    \end{overpic}
  \end{subfigure}%
  \\[4pt]
  \begin{subfigure}{.23\textwidth}
    \begin{overpic}[width=1\linewidth,keepaspectratio]{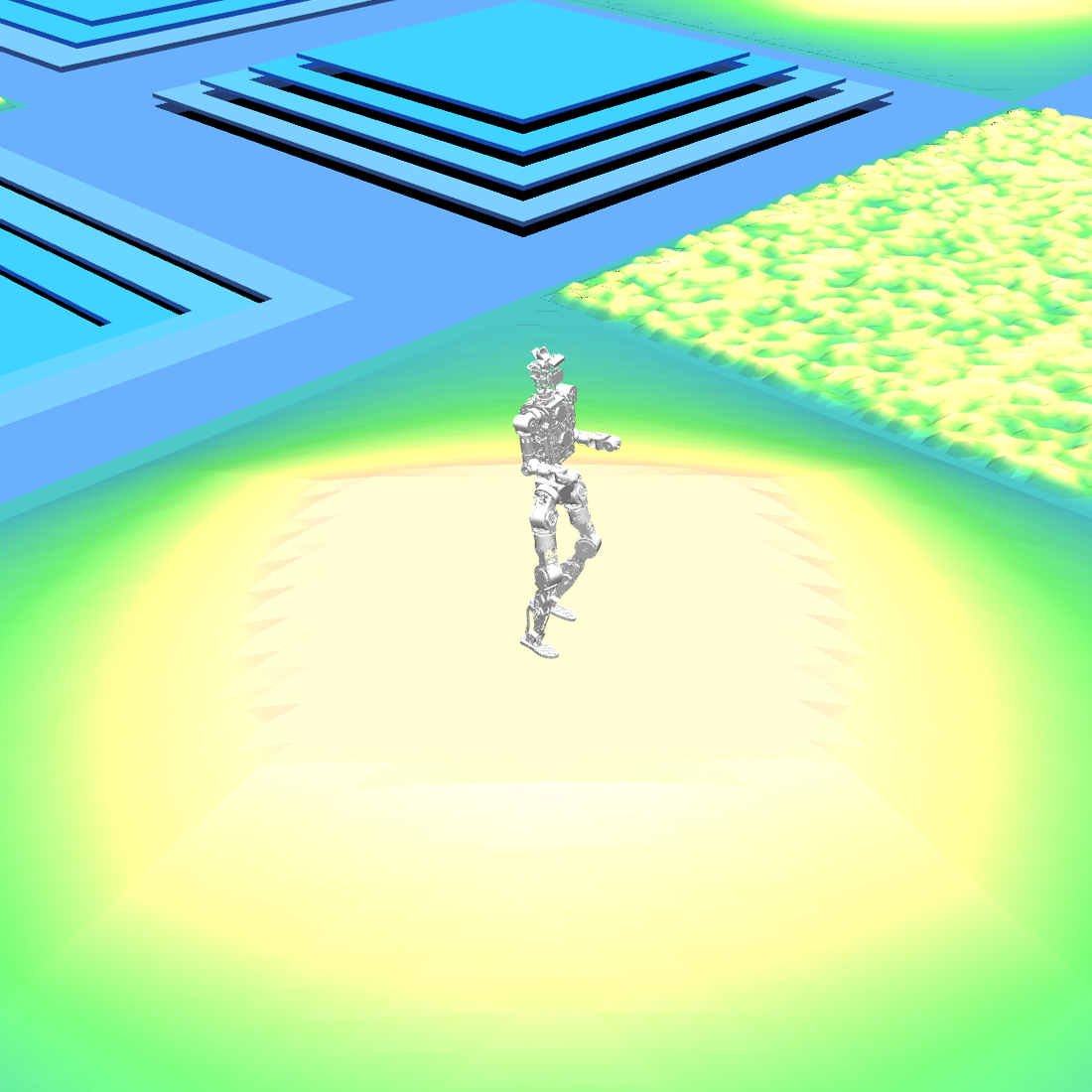}
      \put(3,3){(e)}
    \end{overpic}
  \end{subfigure}\hfill
  \begin{subfigure}{.23\textwidth}
    \begin{overpic}[width=1\linewidth,keepaspectratio]{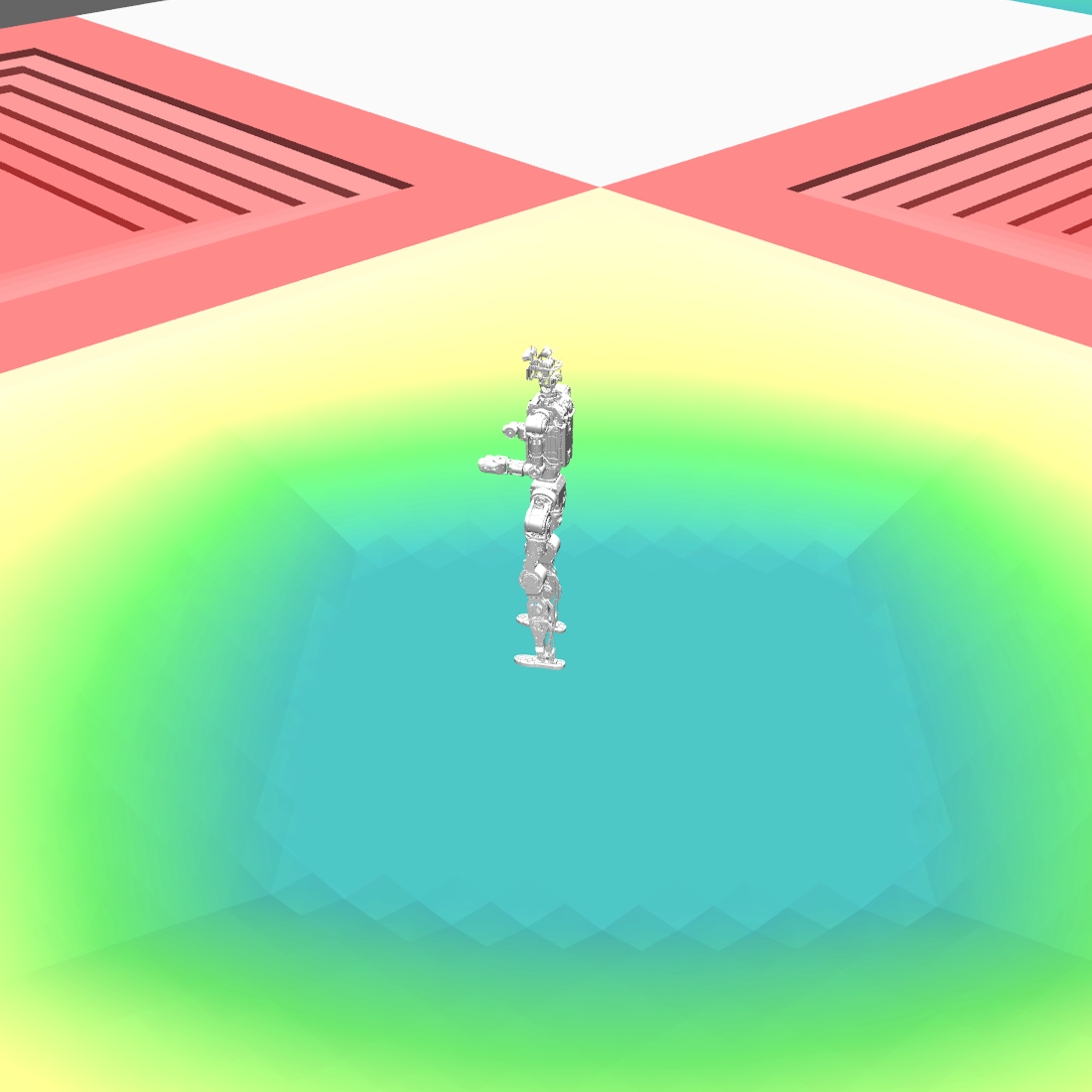}
      \put(3,3){(f)}
    \end{overpic}
  \end{subfigure}\hfill
  \begin{subfigure}{.23\textwidth}
    \begin{overpic}[width=1\linewidth,keepaspectratio]{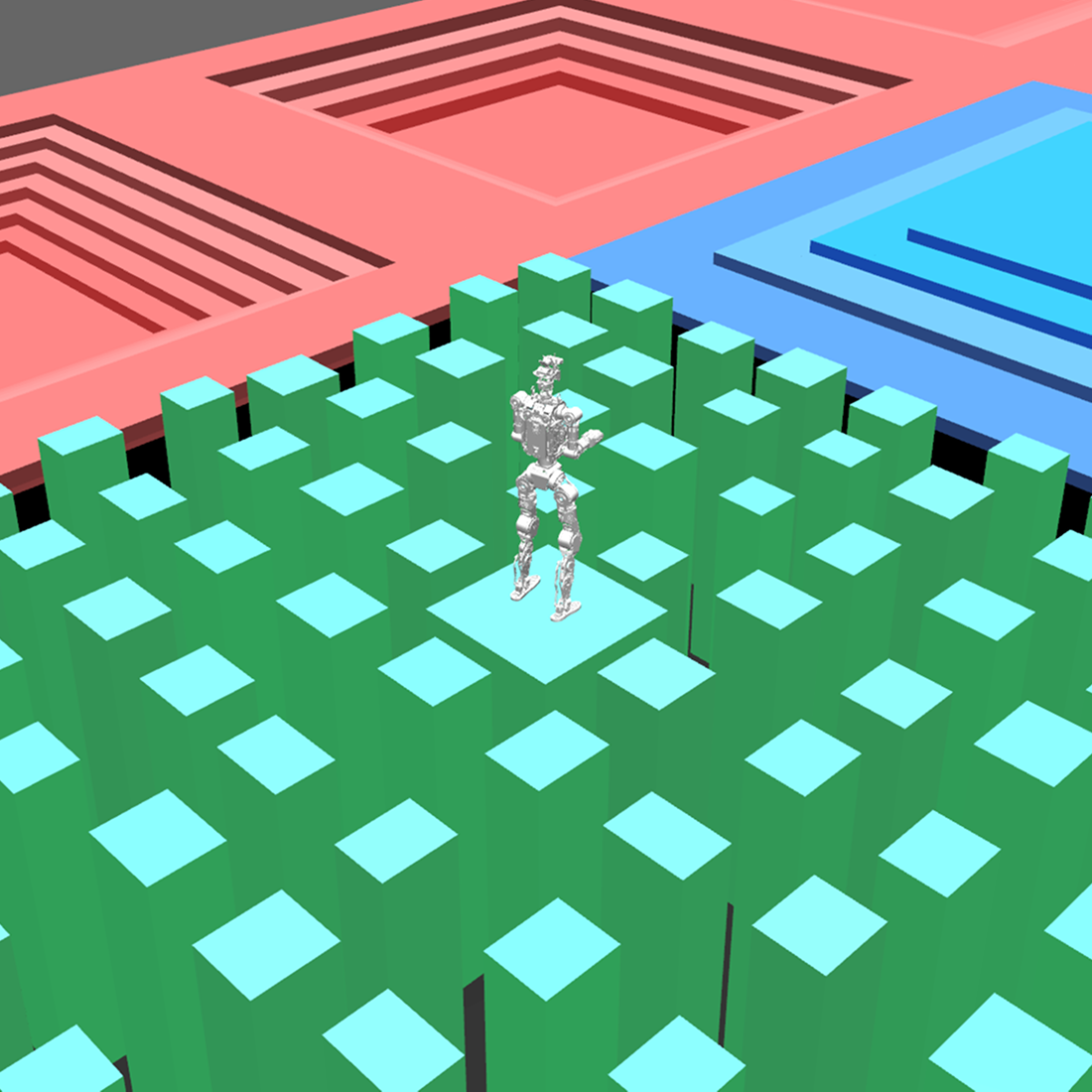}
      \put(3,3){(g)}
    \end{overpic}
  \end{subfigure}\hfill
  \begin{subfigure}{.23\textwidth}
    \begin{overpic}[width=1\linewidth,keepaspectratio]{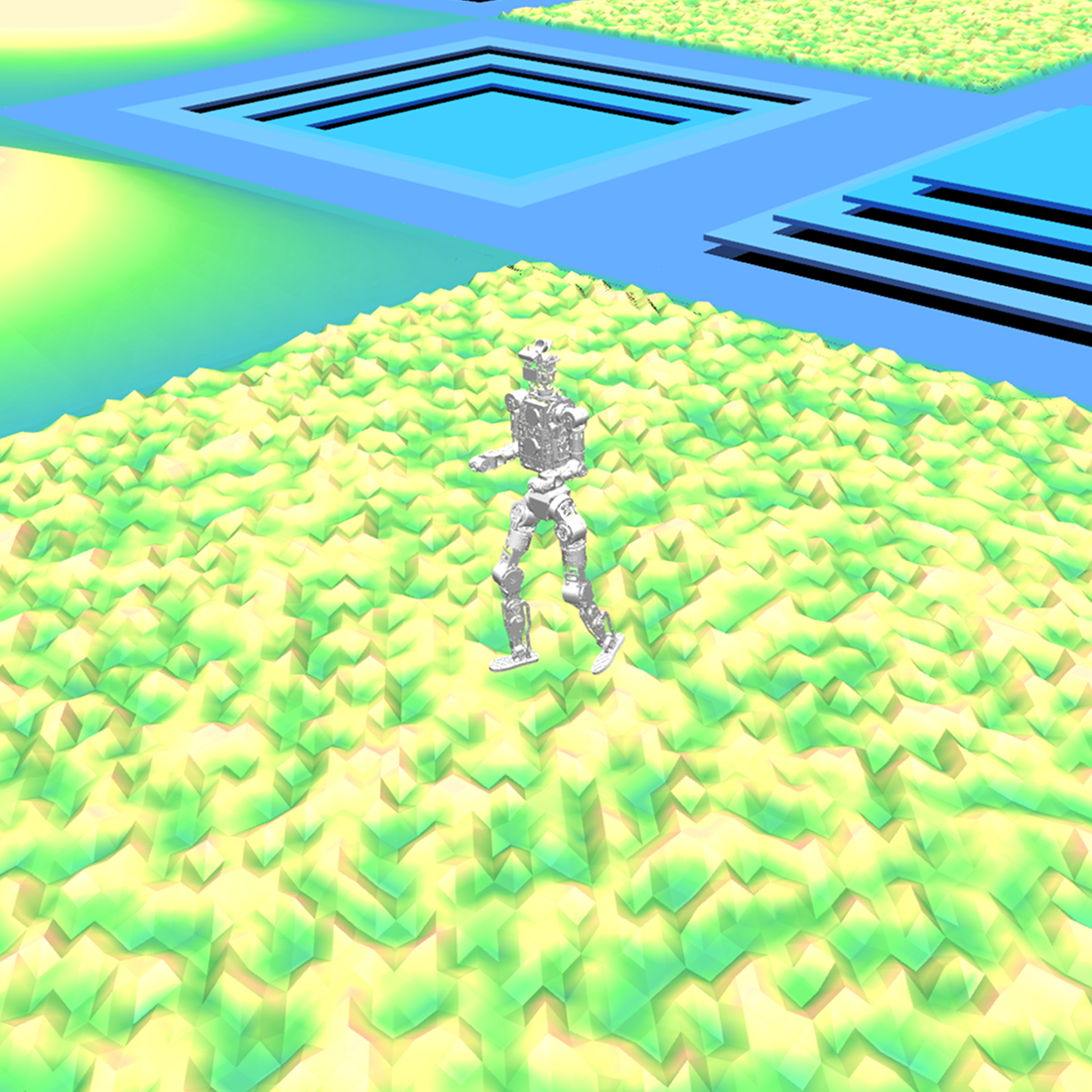}
      \put(3,3){(h)}
    \end{overpic}
  \end{subfigure}%
  \caption{\textbf{Training terrain configurations.}
    (a)~Pyramid stairs, ascending.\@ (b)~Pyramid stairs, descending.
    (c)~Open-width stairs, ascending (no side walls).\@ (d)~Open-width stairs, descending.
    (e)~Pyramid slope, ascending.\@ (f)~Pyramid slope, descending.
    (g)~Stepping stones.\@ (h)~Gravel (random rough height field).
    Stair risers span \SIrange{0.05}{0.20}{\metre} with treads
    \SIrange{0.25}{0.55}{\metre}; slopes up to $23^{\circ}$; rough
    field height $0$--$0.15$\,m. Hard-terrain evaluation extends risers
    to $0.30$\,m (out-of-distribution).}%
  \label{fig:training_terrains}
\end{figure}

\paragraph{Training terrain configurations.}
\Cref{fig:training_terrains} shows the eight terrain types used during
curriculum training. Stair variants (a--d) are the primary training signal
for the multi-channel terrain cost and foothold reward; slope and rough
terrain (e--h) provide gradient and contact-stability diversity. The policy
is never trained with riser heights above $0.20$\,m; the hard-terrain
evaluation extends risers to $0.30$\,m and is fully out-of-distribution.

\paragraph{Evaluation setup.}
All ablation variants are evaluated at iteration 20\,k with 4096 environments
using different random seeds in MuJoCo on stairs, slopes, and rough terrain.
Standard evaluation terrain spans the same riser range as training: stair
flights with risers \SIrange{0.05}{0.20}{\metre} and treads
\SIrange{0.25}{0.55}{\metre}, including 3-step and 5-step flights, open-width
stairs (no side walls), slopes up to $23^{\circ}$, and random rough height
fields $0$--$0.15$\,m. Hard-terrain evaluation uses riser heights
$0.20$--$0.30$\,m with configurations unseen during training. An episode
succeeds if the robot traverses $D_{\mathrm{target}} = \SI{10}{\metre}$ without
triggering fall termination (base tilt $>\ang{70}$, self-collision, or
joint-limit violation). $\mathrm{SR}_\mathrm{hard}$ is computed over the strict
interior riser range $[0.22,\,0.28]\,\mathrm{m}$, excluding the boundary cases
at the edges of the hard sweep.

\paragraph{Ablation variants.}
All variants share the base training setup; only cost weights, reward terms,
and observation inputs differ. Each is evaluated at iteration 20\,k with 4096
environments over different random seeds.

\emph{TACT\,+\,Adaptive Gait (Ours)}: full system with all four terrain-cost
channels ($\alpha_E,\alpha_Q,\alpha_M,\alpha_{\mathrm{climb}}>0$; $\beta>1$
controls lateral asymmetry in $d_{\mathrm{pos}}$), foothold tracking reward,
and gait-adaptive frequency head~\citep{song2025gaitadaptive}.

\emph{TACT-only}: all terrain-cost channels and foothold tracking reward active;
gait-adaptive frequency head removed.

\emph{Adaptive Gait only}: gait-adaptive frequency head added; all terrain-cost
channel weights and foothold tracking reward set to zero
($\alpha_Q{=}\alpha_E{=}\alpha_M{=}\alpha_{\mathrm{climb}}{=}0$,
$w_{\mathrm{foothold}}{=}0$); no privileged height-map input to the critic.

\emph{Baseline}: standard depth-map perceptive policy; neither TACT channels
nor gait adaptation; no privileged critic input.

\paragraph{Payload generalization (terrain).}
Three attachment conditions are evaluated: pelvis-mounted $+$15\,kg and
$+$20\,kg (CoM offset; tests balance under downward wrench) and wrist-mounted
$+$10\,kg (large moment arm; tests trunk-tilt compliance). Mass is added at
test time with no policy modification. A no-mass-DR baseline is trained
identically to the full policy but without waist-link mass offset
randomization. Each condition uses 100 episodes at
$v_x=\SI{0.5}{\metre\per\second}$ on a 5-step staircase with \SI{0.20}{\metre}
risers.

\paragraph{Planned evaluations (future work).}
The following targeted protocols are specified for completeness; their
quantitative results are deferred to future work and not claimed here:
non-nominal pelvis-height tracking
($h^*\in\{h_0{-}6,\,h_0{-}3,\,h_0{+}3\}\,\mathrm{cm}$, recording $h^*$-RMSE),
and a maximum sustained downward pull sweep ($F_{\max}$ at
$\mathrm{SR}>90\,\%$).

\paragraph{Metric definitions.}\label{app:metrics}

\begin{itemize}[leftmargin=1.25em,itemsep=1pt,topsep=2pt]
  \item \textbf{$\mathrm{SR}$}: success rate (\%), fraction of episodes that traverse $D_\mathrm{target}$.
  \item \textbf{$\mathrm{SR}_\mathrm{hard}$}: $\mathrm{SR}$ restricted to riser heights
        $\Delta z\in[0.22,0.28]$\,m; isolates performance on the most demanding
        subset of the hard sweep (strict interior, excluding boundary cases at 0.20 and 0.30\,m).
  \item \textbf{$Q_\mathrm{c}$}: mean flatness cost $Q_i$ (Eq.~\eqref{eq:flatness}) at the
        elevation-map cell under each foot at touchdown, averaged over all touchdown events.
  \item \textbf{$F^{95}$}: $\mathrm{pct}_{95}(\|F_f\|/(mg/2))$, 95th-percentile normalized
        ground reaction force at touchdown events.
  \item \textbf{$E_v$}: velocity-tracking RMSE, $\sqrt{N^{-1}\!\sum_{t=1}^{N}\|v_{xy,t}-v_{\mathrm{cmd},t}\|^2}$\,(m/s).
  \item \textbf{$P$}: mean absolute mechanical power,
        $N^{-1}\!\sum_{t=1}^{N}\|\boldsymbol{\tau}_t\!\odot\!\dot{\mathbf{q}}_t\|_1$\,(W);
        lower values indicate more efficient gait~\citep{sun2026nowyousee}.
  \item \textbf{Foot-target distance}: mean planar Euclidean distance between the
        executed touchdown centroid and the nearest DCM-planned foothold at swing
        initiation, averaged over all touchdown events in the episode (the
        ``foot-target distance'' reported in \cref{sec:traversal}).
\end{itemize}

\subsection{Hardware and Deployment}%
\label{app:hardware}

\paragraph{Sensor architecture.}\label{app:sensors}

The actor policy receives a single perceptual stream at runtime: a noisy depth
image from a forward-facing RGB-D (color-and-depth) camera. A raycaster-based
elevation map mounted at the pelvis is available during training only and is
used exclusively by the privileged critic; it is not available at deployment.
Additional per-foot proximity rays and a pelvis height sensor supply scalar
terrain signals used exclusively in reward computation and privileged critic
observations. The simulation models each sensor with geometry-matched
raycasters; the depth camera adopts the intrinsics of a RealSense D435 and
injects pose domain randomization to bridge the sim-to-real gap.

\paragraph{Depth camera processing pipeline.}\label{app:cam}
Raw depth frames are captured at $30\,\mathrm{Hz}$, cropped to remove border
artifacts, resized to $30{\times}30$\,px, Gaussian-blurred to suppress
raycaster aliasing, and normalized to $[0,1]$. Four temporally spaced frames
are stacked into a history buffer and fed to the depth encoder at each
$50\,\mathrm{Hz}$ control step. Camera intrinsics are matched to a RealSense
D435; extrinsic pose domain randomization (\cref{tab:dr}) covers mounting
uncertainty. The identical pipeline runs at deployment with no modification.

\paragraph{Deployment.}\label{app:deploy}
The trained actor (depth encoder + policy MLP) is exported to ONNX and executed
at $50\,\mathrm{Hz}$ on the robot's onboard computer via a joint-position PD
interface. No real-world fine-tuning is performed; the sim-to-real gap is
closed entirely by domain randomization during training.

\subsection{Additional Results}%
\label{app:supplement}

\subsubsection*{Flat-Terrain Payload Generalization}%
\label{supp:payload_flat}

\begin{minipage}[t]{0.5\textwidth}
  \vspace{0pt}
  Fig.~\ref{fig:payload_w_flat_terrain} evaluates payload generalization across
  three conditions (pelvis $+$15\,kg, pelvis $+$25\,kg, wrist $+$15\,kg) on flat
  terrain and compares Ours against the Baseline, isolating compliance behavior
  from terrain traversal difficulty. At moderate load (pelvis $+$15\,kg, wrist
  $+$15\,kg), Ours maintains 76--79\,\% SR against the Baseline's 67--76\,\%
  while consuming 9--20\,\% less power, consistent with compliance training
  suppressing impulsive GRF recovery. At high centered load (pelvis $+$25\,kg),
  Ours drops to 47.5\,\% (below the Baseline's 60.0\,\%), indicating that a large
  CoM shift exceeds the compliance training distribution and causes
  over-compensation; power remains lower (257 vs.\ 283\,W), confirming the policy
  still reduces impulsive contacts but can no longer maintain balance.
  Wrist-mounted mass, which generates a large moment rather than a direct CoM
  offset, is handled comparably in SR but with a clear power advantage for Ours
  (186 vs.\ 231\,W), consistent with the arm-extended wrench samples in the
  disturbance force-field.
\end{minipage}\hfill
\begin{minipage}[t]{0.47\textwidth}
  \vspace{0pt}
  \centering
  \includegraphics[width=\linewidth]{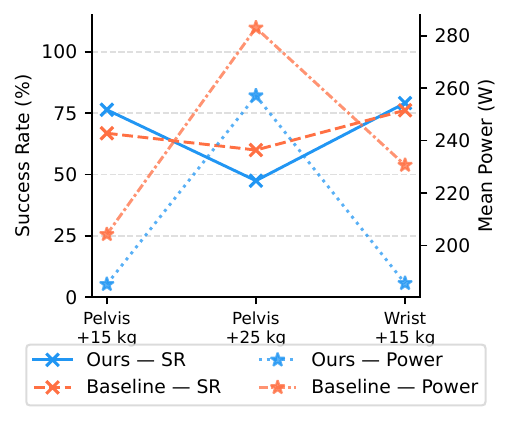}
  \captionof{figure}{SR (\%) and mean power (W) vs.\ payload mass on flat terrain.}%
  \label{fig:payload_w_flat_terrain}
\end{minipage}

\subsubsection*{Cross-Embodiment Generalization}%
\label{supp:cross_embodiment}

\begin{figure}[h!]
  \centering
  \includegraphics[width=\textwidth]{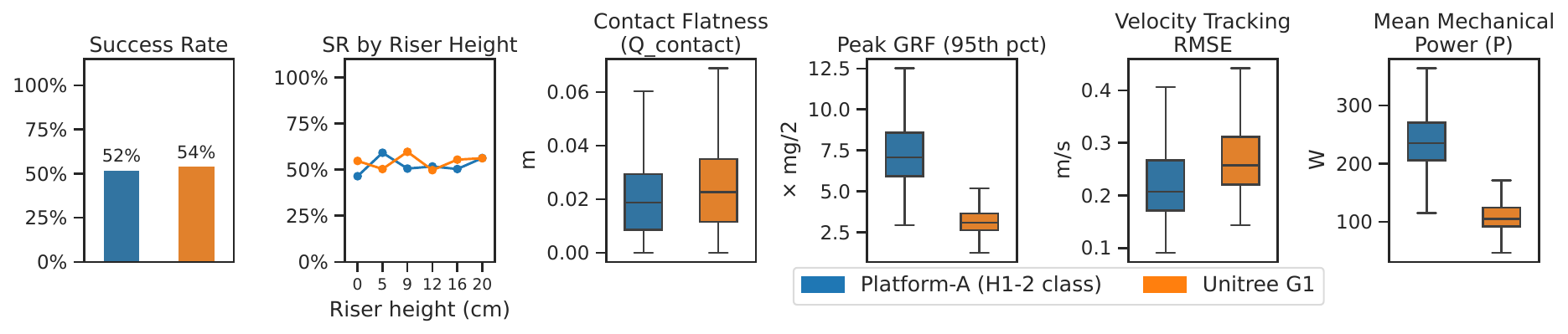}\\[4pt]
  \includegraphics[width=\textwidth]{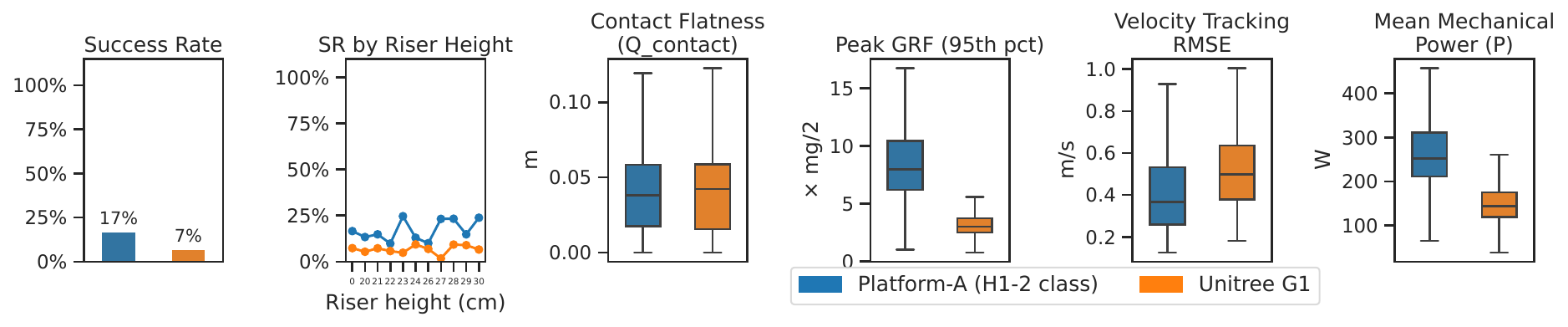}
  \caption{Cross-embodiment ablation on Platform-A (H1-2 class) and Unitree G1:
    \textbf{standard terrain} (top) and \textbf{hard terrain, out-of-distribution}
    (bottom). Identical reward weights and hyperparameters; only robot asset and
    joint configuration differ.}%
  \label{fig:ablation_cross_embodiment}
\end{figure}

Fig.~\ref{fig:ablation_cross_embodiment} shows the ablation evaluated on two
platforms (Platform-A (H1-2 class) and Unitree G1) using identical reward
weights and hyperparameters, with only the robot asset and joint configuration
changed. The relative ordering of variants is preserved across embodiments,
supporting the platform-agnostic claim in \cref{sec:conclusion}.

\subsection{Extended Limitations}%
\label{app:ext_limitations}

\paragraph{Adaptive-gait fragility under kinematic stress.}
The \emph{Adaptive Gait only} variant falls \emph{below} the Baseline on hard
terrain (18\,\% vs.\ 19\,\% SR): EMA-smoothed frequency re-timing without
terrain-quality guidance modifies stance-to-swing departure timing with no
corresponding adjustment to the landing target, producing foot-target
mismatches that the foothold reward would otherwise suppress. Adaptive gait
frequency is therefore an auxiliary capability contingent on accurate foothold
guidance, not an independent locomotion primitive, and its benefit is confined
to regimes where kinematic margins are non-trivial. A natural remedy is to
co-condition the frequency head on the terrain-cost signal so that re-timing
decisions are aware of contact quality, or to gate frequency updates on a
kinematic feasibility check before committing to the new landing target.

\paragraph{Terrain-cost channel redundancy.}
Removing any single terrain cost channel leaves foot-target distance within
5\,\% of the full system ($0.085$--$0.093\,\mathrm{m}$ vs.\
$0.088\,\mathrm{m}$), while removing all three raises it $2.8\!\times$ to
$0.251\,\mathrm{m}$. The channels are thus collectively necessary but
individually non-critical, and the manually tuned weighting ($\alpha_Q{=}4.0$,
$\alpha_E{=}0.6$, $\alpha_M{=}6.0$) does not follow from principled minimality.
A compressed two-channel formulation and a weight-sensitivity sweep across
platforms are left to future work. More broadly, replacing hand-tuned weights
with a differentiable meta-learning or Bayesian optimization pass over the cost
coefficients could yield principled, platform-adaptive weightings without
requiring per-robot re-tuning.

\paragraph{Compliance distribution and upper-body coupling.}
Near 20\,kg of centered CoM shift, two factors compound. First, the
spring-damper sampling concentrates on body-attached downward-biased and
arm-extended forward-biased scenarios; lateral carry, two-point distributed
loads, and dynamically swinging payloads produce wrenches outside this sampled
space. Second, the policy controls leg joints only: at high payload masses,
upper-body inertia shifts the effective CoM in ways the lower-body policy
cannot observe, as the pelvis inertial measurement unit (IMU) and joint
encoders provide no direct signal of trunk angular momentum. Together these set
the effective payload ceiling. Both failure modes have tractable remedies:
broadening the wrench distribution to include lateral and multi-point loads
during training, and extending the controlled joints to include the torso or
arm would directly address the sampling gap and the observability gap
respectively. An online payload estimator feeding a residual compliance target
could further extend the effective range without requiring a larger wrench
training distribution.

\paragraph{Tangent-guided orientation failure on slopes.}
The Bézier swing reference orients the sole using the arc tangent of the
trajectory at the landing point, which on discrete stair steps closely
approximates the riser-face normal and drives the foot to land with the sole
parallel to the tread. On smooth slopes this mechanism fails: the swing
trajectory approaches the inclined surface from above along a path whose
touchdown tangent is approximately horizontal, independent of terrain slope
angle. The resulting orientation reference does not rotate the sole to match
the surface normal, so the foot arrives level on a tilted surface. On ascending
grades the surface rises toward the direction of travel, placing the toe
geometrically closer to the ground than the heel at the moment of touchdown;
the foot therefore contacts toe-first, concentrating GRF on the toe edge. On
descending grades the inverse holds, producing heel-first contact with the toe
elevated. Both deviate from the designed flat-sole landing, narrow the friction
cone, and raise $F^{95}$; on grades above roughly $15^{\circ}$ the resulting
ankle-roll moment is frequently sufficient to trigger fall termination.

A complementary failure arises in the foothold planner. The flatness cost $Q_i$
measures height variance within a local neighbourhood around each candidate
cell: a uniformly inclined surface produces near-zero inter-cell variance, so
$Q_i \approx 0$ regardless of slope angle, and the planner assigns high contact
quality to steep cells. The steepness cost $E_i$ is intended to compensate, but
with the current weighting ($\alpha_E = 0.6$ vs.\ $\alpha_M = 6.0$) it is the
weakest channel and is insufficient to steer footholds away from high-gradient
cells when flatness and reachability are simultaneously satisfied. The
combination of incorrect sole orientation and inadequate slope-avoidance
pressure jointly accounts for the observed degradation on slope terrain
relative to stairs of comparable height gain per step. A corrected planner
would supplement $Q_i$ with a surface-normal tilt term and extend the swing
orientation reference from the trajectory tangent to the estimated surface
normal at the target cell, analogous to normal-aligned foot placement used in
model-based planners.

Additional Bézier-specific failures emerge on slopes. The apex height is set
relative to the take-off elevation; on ascending grades the landing point is
higher than take-off by $\Delta z = l_\text{stride}\sin\theta$, reducing
effective clearance over the landing surface by the same amount. At steep
grades and long strides the foot can contact the slope during mid-swing before
the apex is reached, a form of early collision absent on flat terrain and
stairs. On descending grades the inverse produces excess apex clearance, which
extends swing duration and can desynchronize foot arrival with the DCM-timed
stance transition. Addressing both the orientation and clearance failures
requires replacing the trajectory-tangent reference with a surface-normal
estimate at the target cell (available from the elevation map) and setting the
apex height relative to the landing elevation rather than the take-off
elevation, a straightforward modification to the existing Bézier
parameterization.

\paragraph{Elevation-map dependency and terrain scope.}
The DCM planner consumes a pelvis-mounted elevation map requiring accurate
extrinsic odometry; localization drift degrades foothold selection
monotonically with no fallback planner, and the forward-facing depth camera
provides no lateral or rearward coverage. Highly deformable surfaces (sand,
gravel, foam) violate the rigid-terrain assumption underlying both the Bézier
clearance computation and the flatness cost $Q_i$, causing systematic
under-clearance, and wet or icy contacts fall outside the domain-randomization
friction range ($\mu\in[0.3,1.6]$). Terrain with moving obstacles is not
modeled. Robustness to localization drift could be improved by incorporating a
fallback stance-hold controller that activates when odometry confidence is low,
and by fusing the depth camera into a local elevation map that does not depend
on global pose. Extending the friction domain randomization range and adding
deformable-terrain simulation (e.g., via compliant contact models in MuJoCo)
are the most direct paths to covering low-friction and soft-surface
deployments. Handling dynamic obstacles would require integrating a
short-horizon obstacle prediction model into the foothold search window.

\end{document}